\RequirePackage[svgnames,table]{xcolor}
\documentclass[11pt,letterpaper,logo]{yalearxiv}

\usepackage{natbib}
\setcitestyle{authoryear,square}
\usepackage[normalem]{ulem}
\usepackage{cancel}
\usepackage{verbatim}
\usepackage{float}
\usepackage{subcaption}
\usepackage{multirow}
\usepackage{empheq}

\input{case_study_macros.tex}
\input{prompt_horizontal/prompt_horizontal_macros.tex}

\newtheorem{theorem}{Theorem}

\newif\ifShowYaleLogo
\ShowYaleLogofalse

\newcommand{\YaleLogoHeight}{20pt}

\newcommand{\AdelaideLogoHeight}{18pt}
\newcommand{\CurtinLogoHeight}{18pt}
\newcommand{\SydneyLogoHeight}{20pt}
\newcommand{\UNTLogoHeight}{18pt}

\newlength{\HeaderLogoBoxHeight}
\newcommand{\HeaderLogo}[2][]{%
  \raisebox{\dimexpr0.5\HeaderLogoBoxHeight-0.5\height\relax}%
    [\HeaderLogoBoxHeight][0pt]{%
      \includegraphics[#1]{#2}%
    }%
}

\renewcommand{\firstpagelogos}{%
  \makebox[\textwidth][s]{%

    \ifShowYaleLogo
      \HeaderLogo[
        height=\YaleLogoHeight,
        keepaspectratio
      ]{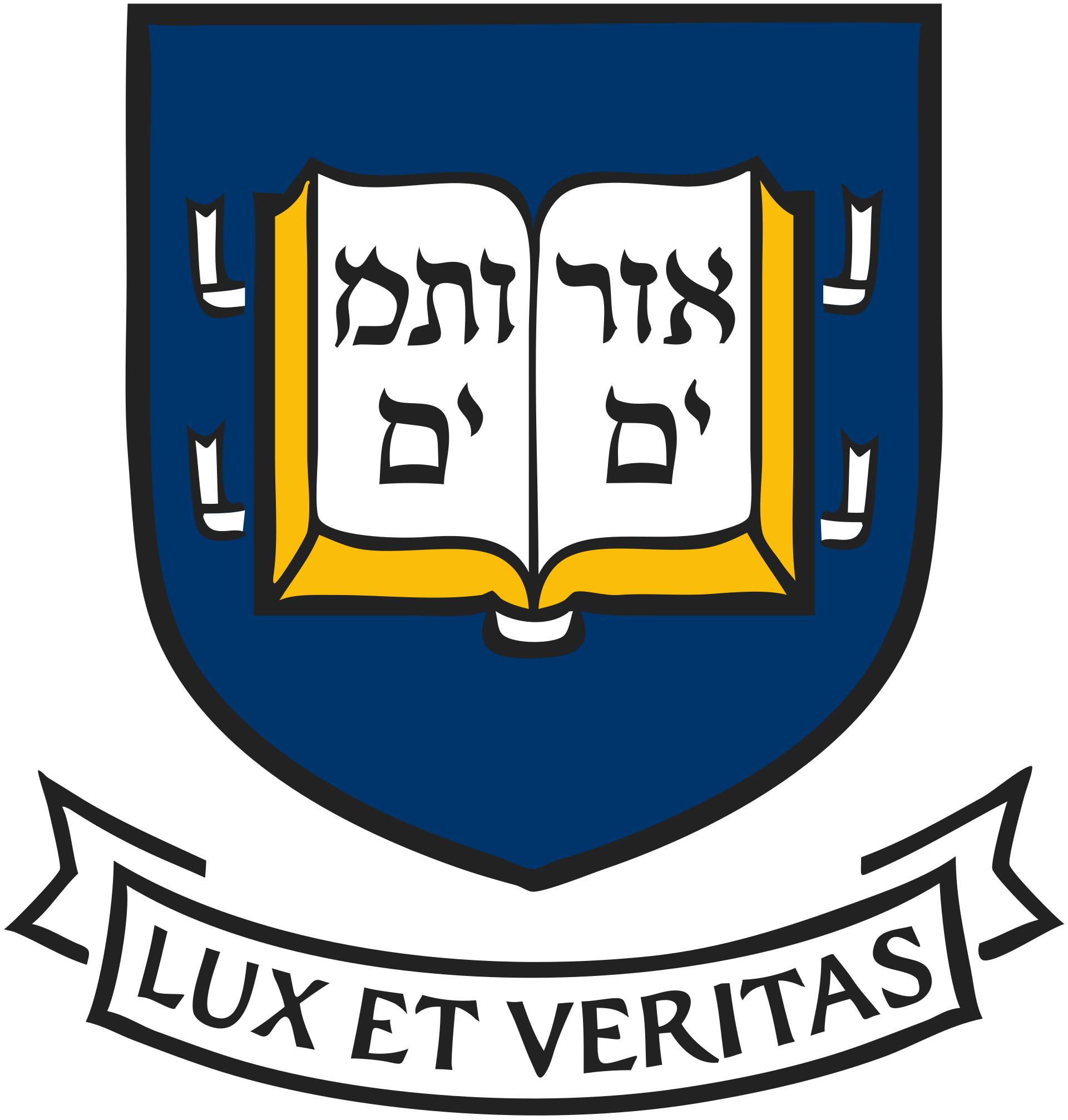}%
      \hfill
    \fi

    \HeaderLogo[
      height=\AdelaideLogoHeight,
      keepaspectratio
    ]{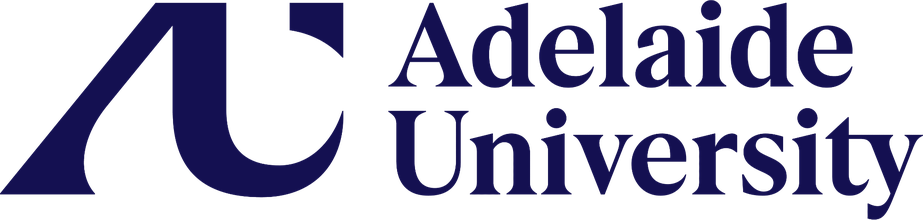}%

    \hfill

        \HeaderLogo[
      height=\SydneyLogoHeight,
      keepaspectratio
    ]{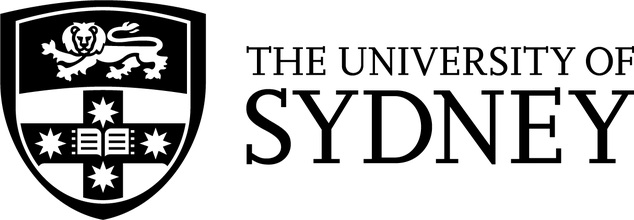}%
\hfill
    \HeaderLogo[
      height=\CurtinLogoHeight,
      keepaspectratio
    ]{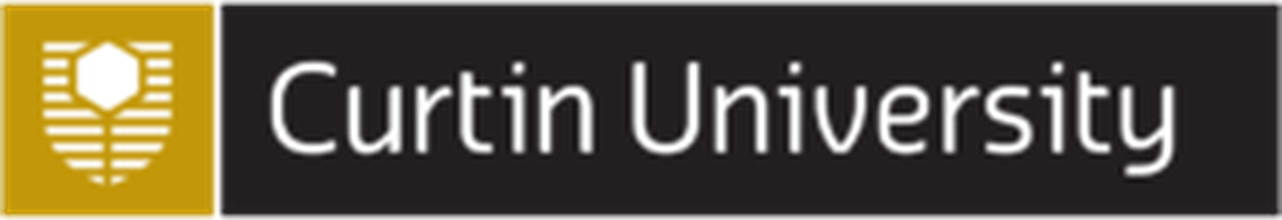}%

    \hfill

    \HeaderLogo[
      height=\UNTLogoHeight,
      keepaspectratio
    ]{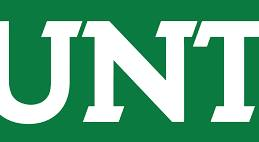}%

  }%
} 

\title{Can We Trust the Teacher? Decoupled Credit Direction-Magnitude for Self-Distillation}
\runningtitle{Can We Trust the Teacher? Decoupled Credit Self-Distillation}
\author{%
Yugu Li$^{1}$, Zehong Cao$^{1,*}$, Peizhen Li$^{1}$, Yang Zhang$^{2}$, Siyi Hu$^{3}$, and Jianglin Qiao$^{4}$\\
$^{1}$School of CSIT, Adelaide University, Adelaide, SA 5000, Australia\\
$^{2}$CAIAA, University of North Texas, Denton, TX 76203, USA\\
$^{3}$School of EECMS, Curtin University, Bentley, WA 6102, Australia\\
$^{4}$ACFR, The University of Sydney, Camperdown, NSW 2050, Australia\\
}
\correspondingauthor{Zehong Cao (\texttt{jimmy.cao@adelaide.edu.au})}

\hypersetup{
  colorlinks=true,
  linkcolor=blue!50!black,
  citecolor=blue!50!black,
  urlcolor=blue!50!black,
  pdftitle={Can We Trust the Teacher? Decoupled Credit Direction-Magnitude for Self-Distillation},
  pdfauthor={Anonymous authors}
}

\begin{document}

\begin{abstract}
{\centering\section*{Abstract}}

RLVR provides reliable trajectory-level credit, while OPSD offers dense supervision for token-level credit. This exposes a fundamental coupling when updating step-level credit direction and magnitude with teacher supervision, preventing steps from receiving reliable credit
directions and contribution magnitudes, while making both vulnerable to teacher judgment errors and preference variance, as supported by our theoretical analysis. To separate credit direction from its contribution magnitude, we introduce \textit{Decoupled Credit Self-Distillation (DCSD)}, which theoretically decouples credit direction and magnitude into two reliable signals and uses them to calibrate privileged teacher supervision. Specifically, we design belief-margin probing to determine credit direction and marginal information gain to quantify credit magnitude, enabling step-to-token credit assignment for policy optimization. Across 11 benchmarks, DCSD achieves the best overall scores against GRPO, OPSD, RLSD, and RLCSD. Compared with base models, DCSD improves the overall score by 8.45 points on mathematical reasoning and 7.01 points on multimodal reasoning, while correcting the credit direction for 6\% of tokens and yielding a 1.5$\times$ reduction in token credit magnitude.


\end{abstract}

\maketitle

\section{Introduction}
\label{intro}

Reinforcement learning with verifiable rewards (RLVR) \citep{lambert2024tulu,shao2024deepseekmath,guo2025deepseek,yu2026dapo} provides reliable outcome feedback, but credit is typically assigned at the trajectory level. Thus, successful trajectories may reinforce incorrect or unnecessary steps, while useful steps in unsuccessful trajectories receive no positive credit. On-policy distillation (OPD) \citep{agarwal2024policy}, particularly on-policy self-distillation (OPSD) \citep{OPSD} and its variants \citep{SDPO,RLSD,pan2026rlcsd,shen2026purified,SRPO}, provides denser supervision by using a privileged teacher to evaluate the student’s reasoning with additional information, such as ground-truth answers or reference solutions.

The key challenge is that a \emph{direct teacher signal may not provide reliable local credit}. Existing OPSD couples credit direction and magnitude through the same teacher signal. The teacher may misjudge whether a step helps or harms the solution, while its signal strength can vary with how privileged information is presented. Consequently, teacher credit can deviate from oracle credit in both \emph{direction} and \emph{magnitude}. RLSD \citep{RLSD} and RLCSD \citep{pan2026rlcsd} improve robustness by anchoring direction to the trajectory-level RL advantage and using self-distillation mainly to modulate update strength. However, this loses local directional flexibility: harmful steps in successful trajectories cannot receive negative credit, nor useful steps in unsuccessful trajectories positive credit. Thus, teacher signals can misdirect, while trajectory anchoring loses local credit. Additional discussion of related work is provided in \textit{Appendix~\ref{related work}}.

Our theoretical insights (see \textit{Section~\ref{sec:reliable_credit}}) reveal a way out of this trade-off. Under an oracle teacher corresponding to the student’s rollout policy conditioned on eventual success, the self-distillation log-ratio is directionally consistent with the oracle local RL advantage, showing that self-distillation contains useful \emph{local directional information}. However, its magnitude does not directly represent a step’s contribution, and a practical teacher can also corrupt its direction. 

\boxed{\text{\parbox{\dimexpr\linewidth-2\fboxsep-2\fboxrule\relax}{
\textbf{Our insights suggest that reliable step-level credit requires separating update direction from contribution magnitude, rather than coupling both in a single teacher signal.}
}}}

\begin{figure}
    \centering
    \includegraphics[width=1.0\linewidth]{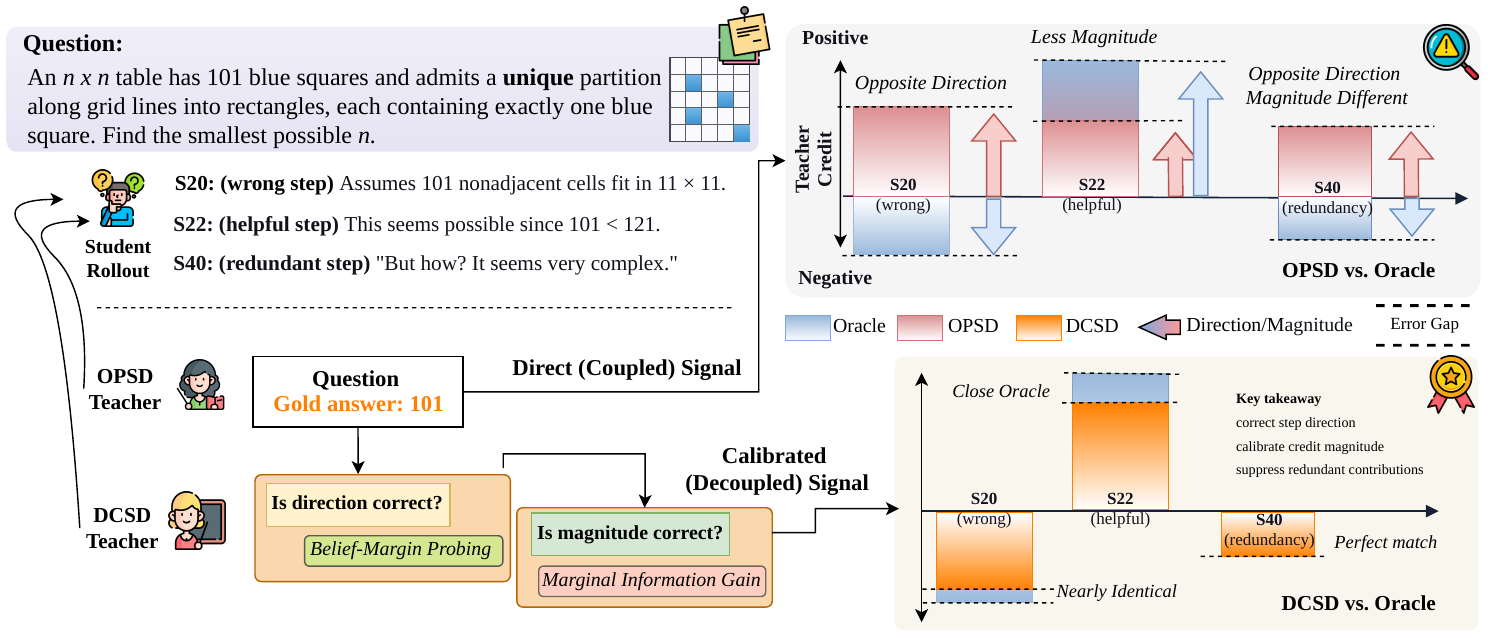}
    \caption{\textbf{Overview of OPSD vs. DCSD credit.}
Direct OPSD can deviate from oracle credit in both direction and magnitude. DCSD decouples these roles to calibrate credit direction and contribution magnitude, reducing mismatch while preserving useful local credit.}
    \label{fig1}
    
\end{figure}

\newtcolorbox{highlightbox}{
    colback=gray!7,
    colframe=gray!7,
    boxrule=0pt,
    arc=0pt,
    boxsep=2pt,
    left=2pt,
    right=2pt,
    top=2pt,
    bottom=2pt,
    hbox
}

\newtcolorbox{insightbox}{
    colback=blue!5,
    colframe=blue!5,
    boxrule=0pt,
    arc=0pt,
    boxsep=0pt,
    left=1pt,
    right=1pt,
    top=2pt,
    bottom=2pt,
    before skip=3pt,
    after skip=3pt
}

Motivated by our theoretical insights, we propose \textbf{Decoupled Credit Self-Distillation (DCSD)} (see \textit{Section~\ref{sec:method}}), which calibrates teacher supervision by separately determining credit direction and contribution magnitude. Specifically, DCSD determines credit direction through \emph{belief-margin probing}, which measures how each reasoning step changes the student’s belief in the correct answer, and quantifies contribution magnitude through \emph{marginal information gain}, which measures how much non-redundant information the step contributes beyond its preceding context. The resulting decoupled credit then calibrates teacher supervision, preserving fine-grained token-level information without allowing teacher variation to freely alter the established step-level direction or magnitude. Thus, DCSD transforms the direct teacher signal in OPSD into a \emph{calibrated teacher signal} that better aligns with oracle step-level credit. \textbf{Figure~\ref{fig1}} provides an illustrative example of credit in OPSD vs. DCSD. The transition from coupled to decoupled credit is formalized as follows:

\noindent
\makebox[\linewidth][c]{%
\begin{minipage}{0.90\linewidth}
\begin{highlightbox}
\makebox[\linewidth][c]{%
{\fontsize{10}{12}\selectfont
$\displaystyle
\mathbf{OPSD:}\;
\underbrace{
[D_k^{T}M_k^{T}]
}_{\rm Coupled\ Credit}
\Rightarrow
\underbrace{
Z_k^{T}
=
Z_k^\star
+
\varepsilon_k^{T}
}_{\rm Direct\ Signal}
\qquad
\mathbf{DCSD:}\;
\underbrace{
[D_k^{T},\,M_k^{T}]
}_{\rm Decoupled\ Credit}
\Rightarrow
\underbrace{
\widetilde Z_k^{T}
}_{\rm Calibrated\ Signal}
\approx
\underbrace{
Z_k^\star
}_{\rm Oracle}
$
}%
}
\end{highlightbox}
\end{minipage}%
}

On the OPSD side, the privileged teacher directly provides the step-level credit $Z_k^T$, whose direction $D_k^T$ and magnitude $M_k^T$ are coupled through the same teacher-derived signal. Consequently, $Z_k^T$ can be viewed as the oracle credit $Z_k^\star$ perturbed by teacher-induced deviation $\varepsilon_k^T$, arising from judgment errors and preference variance. On the DCSD side, we decouple these two components: $D_k^{T}$ is determined by our belief-margin probe, while $M_k^{T}$ is quantified by marginal information gain. Together, they calibrate the teacher signal $\widetilde Z_k^T$ separately toward the oracle credit $Z_k^\star$. 

\textbf{Our contributions are threefold.} \textbf{(1)} We formalize oracle step credit in self-distillation, establish its directional consistency with the local RL advantage, and theoretically identify the factors that cause the direct teacher signal in OPSD to deviate from oracle credit. \textbf{(2)} We introduce DCSD, a decoupled credit framework that calibrates the teacher signal by separately determining credit direction through belief-margin probing and contribution magnitude through marginal information gain, with theoretical guarantees. \textbf{(3)} Experiments across mathematical and multimodal reasoning benchmarks show that DCSD achieves the best overall scores against GRPO, OPSD, RLSD, and RLCSD, demonstrating that calibrated teacher signals improve student overall performance.

\section{Why Direct Teacher Signals Produce Coupled Credit}
\label{sec:reliable_credit}
To understand why direct teacher supervision in OPSD couples credit direction and contribution magnitude, we first formalize the oracle (ground-truth) local credit for each reasoning step, then derive the self-distillation credit that is directionally consistent with the oracle and show how direct teacher supervision deviates from this oracle credit through judgment errors and preference variance.

\subsection{Oracle Local Advantage}
\label{subsec:oracle_credit}
Given an input $x$, let $\pi$ denote the student policy and $\tau=(y_1,\ldots,y_T)\sim\pi(\cdot\mid x)$ a generated response, where $t\in\{1,\ldots,T\}$ indexes response tokens. We partition $\tau$ into $K$ contiguous reasoning steps using boundaries $1=b_0<\cdots<b_K=T+1$, with $k\in\{0,\ldots,K-1\}$ and $C_k=(y_{b_k},\ldots,y_{b_{k+1}-1})$. The state before step $C_k$ is $S_k=(x,C_{<k})$, where $C_{<k}=(C_0,\ldots,C_{k-1})$, and generating $C_k$ yields $S_{k+1}$. We use $\pi(C\mid S)$ to denote the complete-step probability induced by the autoregressive student policy. A terminal verifier provides reward $R(\tau)\in\{0,1\}$ and the underlying on-policy objective computes a trajectory-level advantage $A_\tau$. Let $V^\pi(S)=P_\pi(R(\tau)=1\mid S)$ denote the probability of eventual success from state $S$, and $Q^\pi(S,C)=P_\pi(R(\tau)=1\mid S,C)$ the corresponding success probability after generating step $C$. For the realized step $C_k$, $Q^\pi(S_k,C_k)=V^\pi(S_{k+1})$, giving the oracle local advantage
\begin{empheq}[box=\colorbox{gray!7}]{equation}
A_k^\star
=
Q^\pi(S_k,C_k)-V^\pi(S_k)
=
V^\pi(S_{k+1})-V^\pi(S_k).
\label{eq:oracle_step_advantage}
\end{empheq}

Its sign and magnitude characterize the direction and strength of the local contribution. However, $A_k^\star$ is inaccessible because intermediate-state success probabilities are unknown; $A_\tau$ alone cannot distinguish local contributions within a rollout. Further details are provided in \textit{Appendix~\ref{app:s2_oracle_credit}}.

\subsection{Oracle Signal Determines the Correct Credit Direction}
\label{subsec:ideal_direction}

We first characterize the oracle signal for determining whether a reasoning step should be encouraged or penalized. Consider the student's policy conditioned on eventual success, $\pi^+(C\mid S):=P_\pi(C\mid S,R(\tau)=1)$. By Bayes' rule, $\pi^+(C\mid S)/\pi(C\mid S)=Q^\pi(S,C)/V^\pi(S)$. This gives the oracle signal for step $C_k$ as $Z_k^\star=\log[\pi^+(C_k\mid S_k)/\pi(C_k\mid S_k)]=\log[Q^\pi(S_k,C_k)/V^\pi(S_k)]$. Combining the monotonicity of $\log(\cdot)$ with the oracle local advantage in \textbf{Eq.~\ref{eq:oracle_step_advantage}} gives
\begin{empheq}[box=\colorbox{gray!7}]{equation}
\operatorname{sign}(Z_k^\star)
=
\operatorname{sign}(A_k^\star).
\label{eq:directional_consistency}
\end{empheq}
Thus, $C_k$ is encouraged when $Z_k^\star>0$ and penalized when $Z_k^\star<0$. Importantly, this consistency is directional only: $|Z_k^\star|$ is not the magnitude of the local advantage $|A_k^\star|$. Further details are provided in \textit{Appendix~\ref{app:s2_success_conditioning}}.

\subsection{Direct Teacher Signal Deviates from the Oracle Signal}
\label{subsec:practical_gap}

In practice, the success-conditioned policy $\pi^+$ is unavailable. OPSD instead uses a privileged teacher $\pi_T$ conditioned on additional information $r$ and a preference realization $u$, yielding $Z_k^T=\log[\pi_T(C_k\mid S_k,r,u)/\pi(C_k\mid S_k)]$. Let $\overline{\pi}_T(C\mid S,r)=\mathbb{E}_{u\sim\mu(\cdot\mid S,r)}[\pi_T(C\mid S,r,u)]$ denote the preference-averaged teacher. Then
\begin{empheq}[box=\colorbox{gray!7}]{equation}
Z_k^T
=
Z_k^\star
+
\varepsilon_k^{\mathrm{judge}}
+
\varepsilon_k^{\mathrm{prefer}},
\label{eq:teacher_bridge_decomposition}
\end{empheq}
where $\varepsilon_k^{\mathrm{judge}}=\log[\overline{\pi}_T(C_k\mid S_k,r)/\pi^+(C_k\mid S_k)]$ captures teacher judgment deviation, and $\varepsilon_k^{\mathrm{prefer}}=\log[\pi_T(C_k\mid S_k,r,u)/\overline{\pi}_T(C_k\mid S_k,r)]$ captures preference-dependent variation. Further details of the derivation are provided in \textit{Appendix~\ref{app:teacher_decomposition}}.
 These deviations can reverse the oracle direction when $Z_k^\star Z_k^T<0$. Thus, direct teacher supervision couples \emph{which direction} a step should be updated with \emph{how much} credit it should receive. 
 
In addition, methods such as RLSD \citep{RLSD} and RLCSD \citep{pan2026rlcsd} mitigate directional errors by anchoring local credit to $A_\tau$, e.g., $A_\tau m_k$ with $m_k\geq0$. However, this fixes all local credit to the trajectory-level sign (see \textit{Appendices~\ref{app:s2_opsd}, \ref{app:s2_rlsd}, and~\ref{app:s2_rlcsd}}): a harmful step in a successful trajectory cannot receive negative credit, while a useful step in an unsuccessful trajectory cannot receive positive credit. \textit{Therefore, our analysis motivates decoupling credit direction from magnitude before incorporating teacher evidence for token-level credit assignment.}

\begin{figure}
    \centering
    \includegraphics[width=1.0\linewidth]{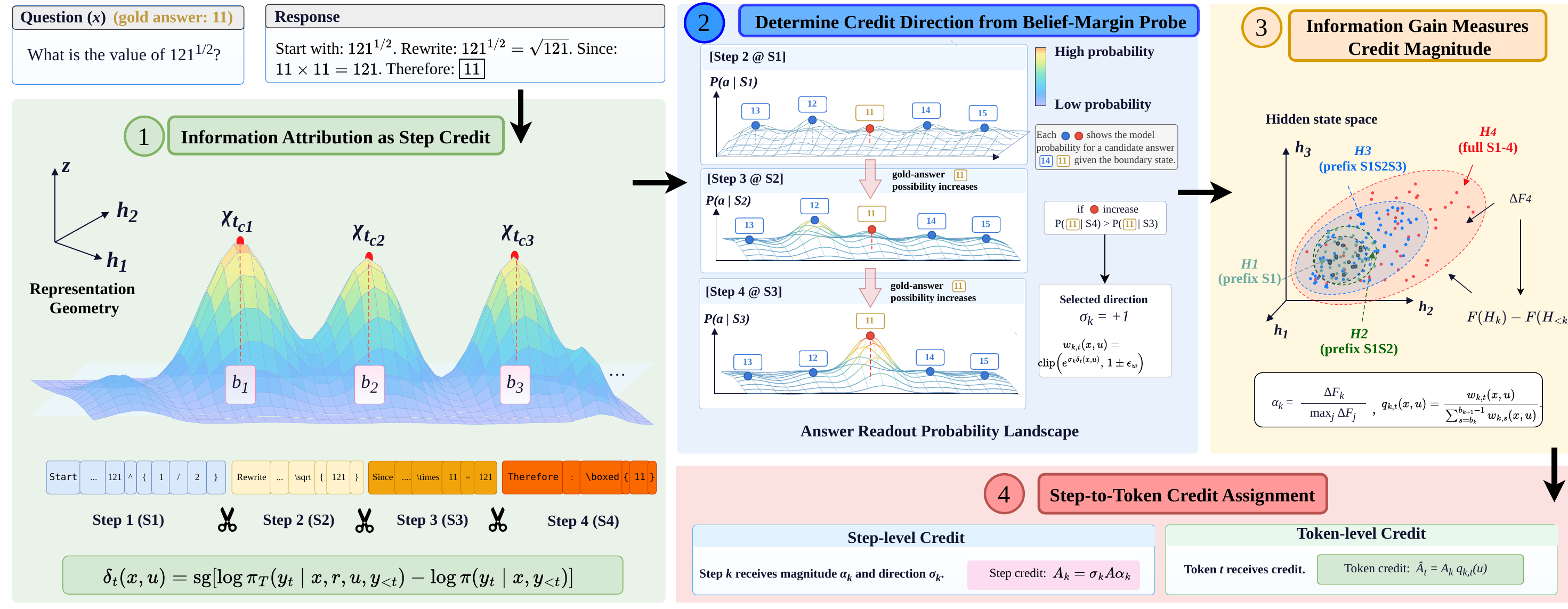}
    \caption{\textbf{DCSD workflow.}
Given a student rollout, DCSD identifies reasoning steps from representation transitions, determines credit direction from answer-belief changes and magnitude from marginal information gain, and uses calibrated teacher signals to distribute step credit across tokens.
}
    \label{fig2}
    
\end{figure}

\section{Method: Decoupled Credit for Calibrating Teacher Signals}
\label{sec:method}

\subsection{From Value Estimation to Information-Attributed Credit}
\label{subsec:information_gain_weighting}

The oracle credit introduced in \textit{Section~\ref{subsec:oracle_credit}} is \emph{value-based}, measuring how each step changes eventual success probability. Estimating it requires intermediate-state values via value estimators or continuation sampling \citep{schulman2017proximal,luo2024improve,setlur2025rewarding}, introducing approximation bias or sampling variance \citep{schulman2015high,kazemnejad2024vineppo,zhang2024rest}. To avoid directly estimating intermediate-state values, DCSD shifts from \emph{value estimation} to \emph{information attribution} as shown in \textbf{Figure~\ref {fig2}} with the full workflow. We characterize each reasoning step using the student’s internal representations, followed by information-theoretic feature attribution \citep{chen2018learning} to quantify the new information it contributes beyond the preceding reasoning context. This provides a measure of \emph{how much} credit a step should receive, while \emph{credit direction} is determined separately through belief-margin probing. The privileged teacher is then used only to distribute the resulting step-level credit across tokens. For autoregressive responses, we adopt marginal representation information as an operational measure of each step's contribution to response construction, using it to assign relative credit magnitude.

\paragraph{Reasoning steps under information attribution.}
Following the view that the structured variation carried by a hidden-state
trajectory can be quantified directly from the representations themselves
\citep{ma2026reasoningmerge}, we let $\mathbf h_t\in\mathbb R^d$ denote the hidden representation of token $y_t$ extracted from a fixed student layer. For each reasoning step $C_k$, let $\mathcal H_k=\{\mathbf h_t\}_{t=b_k}^{b_{k+1}-1}$ denote its hidden-state feature group, with $\mathcal H_{<k}=(\mathcal H_0,\ldots,\mathcal H_{k-1})$ and $\mathcal H_{\leq k}=(\mathcal H_0,\ldots,\mathcal H_k)$. Let $X$ and $Y$ denote the random input and response corresponding to $x$ and $\tau$, respectively. We identify the step boundaries from representational transitions along the generated response using change-point detection; the detailed construction is provided in \textit{Appendix~\ref{app:information_attribution_details}}. For a target $\Upsilon$ and ordered feature groups $\Phi_k$
with finite total information, the mutual-information chain
rule yields the following sequential attribution.

\newtcolorbox{theorembox}{
    colback=gray!3,
    colframe=gray!35,
    boxrule=0.5pt,
    arc=1pt,
    left=2pt,
    right=2pt,
    top=2pt,
    bottom=2pt,
    before skip=1pt,
    after skip=1pt
}

\vspace{1mm}

\begin{theorembox}
\begin{theorem}[Sequential Information Attribution]
\label{thm:sequential_information_attribution}
Define $\mathcal I_k^{\Upsilon}=I(\Upsilon;\Phi_k\mid X,\Phi_{<k})$. Then $\mathcal I_k^{\Upsilon}=I(\Upsilon;\Phi_{\le k}\mid X)-I(\Upsilon;\Phi_{<k}\mid X)\ge0$ and $\sum_{k=0}^{K-1}\mathcal I_k^{\Upsilon}=I(\Upsilon;\Phi_{0:K-1}\mid X)$. For $\Upsilon=Y$ and $\Phi_k=\mathcal H_k$, $\mathcal I_k=\mathcal I_k^{Y}$ attributes response information sequentially to steps without assigning correctness labels.
\end{theorem}
\end{theorembox}

\vspace{1mm}

The proof of Theorem~\ref{thm:sequential_information_attribution} is provided in \textit{Appendix~\ref{app:proof_information_attribution}}.



\subsection{Belief-Margin Probe Design for Decoupled Credit Direction}
\label{subsec:belief_probe}

We determine step-level credit direction from the student's own answer belief, independently of privileged teacher feedback. At each state $S_k$, we append the same fixed answer-readout suffix $\mathcal P_{\mathrm{ans}}$ and score complete candidate answers without sampling additional reasoning. Let $a^+$ denote the correct answer and $\mathcal A_\tau$ a fixed candidate set containing $a^+$, the rollout prediction, and valid candidates collected across step boundaries, with at least one competitor. For each $a\in\mathcal A_\tau$, define the complete-answer log-likelihood as $\ell_k(a)=\log\pi(a\mid S_k,\mathcal P_{\mathrm{ans}})$. Candidate construction, answer serialization, and sequence scoring are detailed in \textit{Appendix~\ref{app:belief_probe_details}}.

\paragraph{Belief-margin probe design.}
At state $S_k$, we measure the student's relative support for the correct answer by
$
M_k
=
\ell_k(a^+)
-
\log\!\left(
\sum_{a\in\mathcal A_\tau\setminus\{a^+\}}
\exp(\ell_k(a))
\right).
$
The step-induced belief change is
\begin{equation}
\Delta M_k=M_{k+1}-M_k.
\label{eq:belief_margin_change}
\end{equation}
Thus, the sign of $\Delta M_k$ indicates whether $C_k$ shifts belief toward or away from the correct answer.

\paragraph{Belief-guided credit direction.}
Because small margin changes may be unreliable due to readout mismatch or incomplete candidate coverage, we use the local direction only when $\Delta M_k$ crosses the corresponding confidence threshold; otherwise, we fall back to the trajectory-level direction:
\begin{equation}
\sigma_k
=
\begin{cases}
\operatorname{sign}(\Delta M_k),
& \Delta M_k\geq\tau_M^+ \ \text{or}\ \Delta M_k\leq\tau_M^-,\\
\operatorname{sign}(A_\tau),
& \text{otherwise},
\end{cases}
\label{eq:step_direction}
\end{equation}
where $\tau_M^-<0<\tau_M^+$ set the error tolerance for Theorem~\ref{thm:forward_readout_reliability}.

\vspace{2mm}

\begin{theorembox}
\begin{theorem}[Oracle-Consistent Credit Direction]
\label{thm:forward_readout_reliability}
Let $D_k^\star$ be the oracle log-odds change over $C_k$ and $\Xi_k$ the probe's readout-and-coverage error. For $(S_k,S_{k+1})$ with $0<V^\pi(S_k),V^\pi(S_{k+1})<1$ and finite $\Xi_k$, $|\Delta M_k-D_k^\star|\le\Xi_k$. Hence, if $\Xi_k<\min\{\tau_M^+,-\tau_M^-\}$, then $\Delta M_k\notin(\tau_M^-,\tau_M^+)$ implies $\sigma_k=\operatorname{sign}(\Delta M_k)=\operatorname{sign}(A_k^\star)=\operatorname{sign}(Z_k^\star)$.
\end{theorem}
\end{theorembox}

\vspace{1mm}

The proof of Theorem~\ref{thm:forward_readout_reliability} is provided in \textit{Appendix~\ref{app:proof_direction}}.


\subsection{Information-Gain Principle for Decoupled Credit Magnitude}
\label{subsec:magnitude_control}

Having determined the credit direction $\sigma_k$, we next quantify \emph{how much} credit each reasoning step should receive using the information-attribution principle introduced in \textit{Section~\ref{subsec:information_gain_weighting}}. This determines credit magnitude independently of privileged teacher supervision.

\paragraph{Marginal information gain.}
For any representation collection $\mathcal S$, define its information volume as $F(\mathcal S)=\frac{1}{2}\log\det\!\left(\mathbf I_d+\beta\sum_{\mathbf h\in\mathcal S}\mathbf h\mathbf h^\top\right)$, where $\mathbf I_d$ is the $d$-dimensional identity matrix and $\beta>0$ is fixed. The marginal information gain of step $C_k$ is
\begin{equation}
\Delta F_k
=
F\!\left(\mathcal H_{\leq k}\right)
-
F\!\left(\mathcal H_{<k}\right).
\label{eq:conditional_information_gain}
\end{equation}
Thus, $\Delta F_k$ measures the information introduced by $C_k$ beyond the preceding reasoning. \textbf{Theorem~\ref{thm:information_gain_magnitude}} shows that it is exactly the sequential attribution of \textbf{Theorem~\ref{thm:sequential_information_attribution}} for a latent answer variable.

\paragraph{Information gain quantifies credit magnitude.}
We convert the marginal information gains into bounded relative credit magnitudes as $\alpha_k=\Delta F_k/\max_{0\leq j<K}\Delta F_j$, where $0\leq\alpha_k\leq1$ and $\max_k\alpha_k=1$, whenever $\max_j\Delta F_j>0$. Thus, $\alpha_k$ measures the relative magnitude of step $C_k$ with respect to the most informative step in the response. More details can be found in \textit{Appendix~\ref{app:magnitude_details}}. 
DCSD assigns relative step-credit magnitude according to marginal information contribution within the response, while Theorem~3 connects the underlying information gain to oracle credit through evidence-model bounds rather than asserting pointwise recovery of $|A_k^\star|$.

\begin{theorembox}
    \begin{theorem}[Information Gain Bounds Credit Magnitude]
\label{thm:information_gain_magnitude}
Under (E1)--(E3) in Appendix~\ref{app:proof_information_magnitude}, $\Delta F_k=I(\Theta;O_{\mathcal H_k}\mid O_{\mathcal H_{<k}})=\sup_f I(f(\Theta);O_{\mathcal H_k}\mid O_{\mathcal H_{<k}})$, over finite-valued measurable answer maps $f$. For $0<V^\pi(S_k)<1$, expectation over step evidence gives $\mathbb E[A_k^\star]=0$, $\mathbb E[(A_k^\star)^2]\le\min\{\frac12\Delta F_k,\frac14\}$, and $\mathbb E[Z_k^\star\mid R=1]\le\Delta F_k/V^\pi(S_k)$. Moreover, $\Delta F_k=0$ iff $A_k^\star=0$ for every $f$. Thus $\Delta F_k$ is the sharp verifier-independent answer-information envelope.
\end{theorem}
\end{theorembox}

\vspace{1mm}

The proof of Theorem~\ref{thm:information_gain_magnitude} is provided in \textit{Appendix~\ref{app:information_value_bridge}}.


\subsection{Calibrated Teacher Supervision for Step-to-Token Credit Assignment}
\label{subsec:step_objective}

The preceding components determine \emph{whether} each reasoning step should be encouraged or penalized through $\sigma_k$, and \emph{how much} credit it should receive through $\alpha_k$. We now use the privileged teacher only to distribute this established step-level credit among tokens within each step. Given privileged content $r$ and preference realization $u$, define the token-level teacher--student discrepancy as $\delta_t=\operatorname{sg}\!\left[\log\pi_T(y_t\mid x,r,u,y_{<t})-\log\pi(y_t\mid x,y_{<t})\right]$, where $\operatorname{sg}$ stops gradients. This signal provides teacher evidence without determining the step-level direction or magnitude.

\paragraph{Teacher-calibrated token credit allocation.}
For $t\in C_k$, we align teacher evidence with the established direction and bound its influence using $w_{k,t}=\operatorname{clip}\!\left(e^{\sigma_k\delta_t},1-\epsilon_w,1+\epsilon_w\right)$, where $0\leq\epsilon_w<1$. We normalize these weights within $C_k$ as $q_{k,t}=w_{k,t}\Big/\sum_{s\in C_k}w_{k,s}$. By construction, $\sum_{t\in C_k}q_{k,t}=1$, so $q_{k,t}$ only determines within-step allocation.

\paragraph{Step-to-token credit assignment.}
Let $\kappa_\tau>0$ be a teacher-independent scale factor. For $t\in C_k$, the token-level credit is
\begin{equation}
A_t^{\mathrm{tok}}
=
\sigma_k\kappa_\tau|A_\tau|\alpha_k q_{k,t},
\qquad
b_k\leq t<b_{k+1}.
\label{eq:dcsd_token_credit}
\end{equation}
This factorization separates step direction $\sigma_k$, step magnitude $\kappa_\tau|A_\tau|\alpha_k$, and within-step teacher allocation $q_{k,t}$. Since $\alpha_k$ is max-normalized rather than sum-normalized, the total absolute response credit is $\kappa_\tau|A_\tau|\sum_k\alpha_k$ and need not equal $\kappa_\tau|A_\tau|$. If $A_\tau=0$, all token credits are zero.  The clipping keeps all teacher weights finite and positive. Optimization details and the scale convention are provided in \textit{Appendix~\ref{app:step_objective_details}}.

\vspace{1mm}

\begin{theorembox}
    \begin{theorem}[Calibrated Teacher Supervision Preserves Step Credit]
\label{thm:teacher_error_isolation}
Fix the student policy, response, partition, $\sigma_k\in\{-1,+1\}$, $\alpha_k\ge0$, $\kappa_\tau>0$, and $A_\tau\ne0$. For every teacher realization, $\sum_{t\in C_k}|A_t^{\mathrm{tok}}|=\kappa_\tau|A_\tau|\alpha_k$ and $\operatorname{sign}(A_t^{\mathrm{tok}})=\sigma_k$ for $t\in C_k$ with $\alpha_k>0$. Hence, on steps certified by Theorem~\ref{thm:forward_readout_reliability}, nonzero token credits carry $\operatorname{sign}(A_k^\star)$, and step credit is proportional, within a response, to $\Delta F_k$ (Theorem~\ref{thm:information_gain_magnitude}).
\end{theorem}
\end{theorembox}

\vspace{1mm}

The proof of Theorem~\ref{thm:teacher_error_isolation} is provided in \textit{Appendix~\ref{app:proof_teacher_isolation}}.


\vspace{3mm}

Taken together, Theorems 1–4 establish the theoretical foundation of DCSD for decoupling credit direction and magnitude in OPSD, while calibrating teacher supervision rather than direct signal. We formalize and expand upon the brief formulation introduced in \textit{Section~\ref{intro}}, as summarized below.

\noindent
\makebox[\linewidth][c]{%
\begin{minipage}{0.90\linewidth}
\begin{highlightbox}
\makebox[\linewidth][c]{%
{\fontsize{8}{10}\selectfont
$\displaystyle
\mathbf{OPSD:}\;
\underbrace{
[\operatorname{sign}(Z_k^T),\,|Z_k^T|]
}_{\text{Coupled Credit}}
\Rightarrow
\underbrace{
Z_k^T
=
Z_k^\star
+
\varepsilon_k^{T}
}_{\text{Direct Signal}}
\qquad
\mathbf{DCSD:}\;
\underbrace{
[\sigma_k,\,\alpha_k]
}_{\text{Decoupled Credit}}
\Rightarrow
\underbrace{
A_t^{\mathrm{tok}}
=
\sigma_k\kappa_\tau |A_\tau|\alpha_k q_{k,t}
}_{\text{Calibrated Credit}}
$
}%
}
\end{highlightbox}
\end{minipage}%
}


\section{Experiments and Results}

\subsection{Experimental Setup}

\paragraph{Training setup.}
We consider two reasoning settings: mathematical and multimodal reasoning. For mathematical reasoning, we use Qwen3-4B~\citep{yang2025qwen3} as the base model and DAPO-17K~\citep{yu2026dapo} for training; for multimodal reasoning, we use Qwen3-VL-8B-Instruct~\citep{bai2025qwen3} with MMFineReason-123K~\citep{lin2026mmfinereason}. Our mathematical runs share the base model, training data, and rollout budget; multimodal baselines retain the published settings of \citet{RLSD}. Training is implemented with verl~\citep{sheng2025hybridflow}; hyperparameter settings are provided in \textit{Appendix~\ref{app:training}}, with hardware specifications and computational costs reported in \textit{Appendix~\ref{computational overhead}}.

\paragraph{Baselines, benchmarks, and metrics.}
We compare DCSD against GRPO~\citep{shao2024deepseekmath}, OPSD~\citep{OPSD}, and RLSD~\citep{RLSD}, with the corresponding base models as references. For mathematical reasoning, we additionally compare with RLCSD~\citep{pan2026rlcsd} separately because it uses a different construction of privileged information. We evaluate mathematical reasoning on AIME24/25/26 \citep{maa_aime_2024,maa_aime_2025,aime26}, AMC23 \citep{maa_amc_2023}, MATH500~\citep{hendrycksmath2021}, and HMMT \citep{dekoninck2026matharena}, and multimodal reasoning on MMMU~\citep{yue2024mmmu}, MathVista~\citep{lu2024mathvista}, MathVision~\citep{wang2024measuring}, ZeroBench-Sub~\citep{roberts2025zerobench}, and WeMath~\citep{qiao2025we}. We report mean@4 accuracy (\%), with Overall computed as the sample-weighted average across benchmarks. For multimodal tasks, we train only DCSD and report baseline results from \textbf{Table~2} of RLSD under the same settings. We follow the same training and evaluation settings for fair comparison, with further details provided in \textit{Appendix~\ref{app:baseline-training}}.

\paragraph{Training dynamics and credit analysis.}
We further examine how different credit designs affect learning by tracking training reward and validation accuracy throughout training. For DCSD, we measure the credit direction-correction rate, including positive and negative corrections, to quantify how often local credit departs from the trajectory-level direction. We also compare the mean absolute RL advantage before and after magnitude allocation to quantify how DCSD adjusts credit magnitude. Definitions and aggregation details are provided in \textit{Appendix~\ref{app:dynamics}}.

\paragraph{Teacher signal reliability: direct vs.\ calibrated.}
To isolate teacher-signal deviation and assess the effect of calibration, we fix the Qwen3-4B responses to 90 AIME24–AIME26 problems and vary only the privileged information provided for scoring. We consider five conditions: no answer (\textit{C0}), the correct answer (\textit{C1}), an incorrect answer (\textit{C2}), an equivalent form of the correct answer (\textit{C3}), and the correct answer with a worked solution (\textit{C4}). Using Qwen3-32B under the same conditions as a stronger-model reference, which we treat as a proxy for the oracle teacher, we quantify two complementary forms of deviation. First, \emph{Top-1 Disagreement Rate (TDR)} captures teacher judgment errors by measuring the percentage of token positions at which the model’s full-vocabulary top-1 prediction differs from the reference. Second, \emph{Mean Absolute Error (MAE)} captures teacher preference variation by measuring the mean absolute difference in sampled-token log-probabilities, reported in $10^{-3}$ nat/token. Comparing TDR and MAE across C0–C4 therefore reveals how teacher judgments and preference strengths deviate under different privileged information, and how effectively each method calibrates these deviations. Further details are provided in \textit{Appendix~\ref{app:diagnostics}}.

\subsection{Main Performance Results}
\label{sec:main_results}

\paragraph{Mathematical reasoning tasks.}
\textbf{Table~\ref{tab:math_reasoning_results}} shows that GRPO and RLSD generally outperform OPSD, suggesting that anchoring credit to outcome feedback is more robust than directly following an unconstrained teacher signal. However, trajectory-level anchoring assigns the same direction to all steps and therefore cannot capture local credit reversals. DCSD consistently improves over these baselines, indicating that robust credit direction and local flexibility need not be traded off: decoupling direction and magnitude allows local credit to be refined without directly inheriting teacher deviations. The gains on AIME24-26 and HMMT further demonstrate the benefit of this design on challenging multi-step reasoning, while improvements on the higher-accuracy AMC23 and MATH500 show fine-grained credit remains useful even when trajectories are frequently successful.

\paragraph{Multimodal reasoning tasks.}
As shown in \textbf{Table~\ref{tab:multimodal_results}}, RLSD is the strongest baseline, indicating the benefit of combining outcome anchoring with fine-grained teacher supervision. DCSD further improves performance on MMMU, MathVista, MathVision, and WeMath, with the largest gain on WeMath. As WeMath contains compositional problems involving multiple knowledge concepts and intermediate sub-problems, this gain is consistent with our motivation for local credit assignment: useful intermediate reasoning can receive positive credit even when the overall trajectory is unsuccessful, while marginal information gain differentiates its contribution strength. This suggests that decoupling direction and magnitude can better preserve and weight useful local reasoning than assigning a common outcome direction to all steps. We notice ZeroBench is the exception where DCSD underperforms RLSD. As these problems place greater demands on fine-grained visual perception and spatial reasoning, errors originating from visual evidence may also limit the subsequent reasoning and local credit estimation, which cannot be addressed through credit calibration alone.

\begin{table}[t]
\centering
\small
\setlength{\tabcolsep}{5pt}
\caption{Reasoning performance (mean@4). Best results are \textbf{bold}.}
\label{tab:main_results}
\begin{subtable}{\linewidth}
\centering
\phantomsubcaption\label{tab:math_reasoning_results}
\phantomsubcaption\label{tab:multimodal_results}
\begin{tabular}{lcccccc|c}
\toprule
Method & \multicolumn{6}{c|}{Benchmarks} & Overall \\
\midrule
\multicolumn{8}{l}{\textit{(a) Mathematical reasoning (Qwen3-4B)}} \\
 & AIME24 & AIME25 & AIME26 & AMC23 & MATH500 & HMMT & \\
\cmidrule(lr){2-7}
Vanilla & 65.83 & 54.17 & 52.50 & 92.50 & 88.10 & 26.67 & 81.40 \\
GRPO    & 67.50 & 55.83 & 50.00 & 90.00 & 92.20 & 33.33 & 84.70 \\
OPSD    & 50.00 & 46.67 & 45.83 & 87.50 & 88.60 & 26.67 & 80.11 \\
RLSD    & 63.33 & 52.50 & 55.83 & 92.50 & 91.00 & 33.33 & 83.86 \\
\rowcolor{blue!10}
DCSD (\textit{Ours}) & \textbf{68.33} & \textbf{57.50} & \textbf{59.17} & \textbf{95.00} & \textbf{96.00} & \textbf{36.67} & \textbf{88.56} \\
\midrule
\multicolumn{8}{l}{\textit{(b) Multimodal reasoning (Qwen3-VL-8B-Instruct)}} \\
 & MMMU & MathVista & MathVision & ZeroBench & WeMath & & \\
\cmidrule(lr){2-6}
Vanilla & 62.44 & 73.80 & 47.37 & 19.76 & 54.10 & & 53.43 \\
GRPO    & 65.11 & 76.20 & 48.82 & 22.60 & 56.57 & & 55.49 \\
OPSD    & 63.82 & 75.10 & 47.53 & 21.06 & 54.95 & & 54.13 \\
RLSD    & 67.22 & 78.10 & 52.73 & \textbf{24.85} & 58.00 & & 58.19 \\
\rowcolor{blue!10}
DCSD (\textit{Ours}) & \textbf{68.44} & \textbf{79.20} & \textbf{54.11} & 20.66 & \textbf{67.05} & & \textbf{61.14} \\
\bottomrule
\end{tabular}
\end{subtable}

\end{table}

\begin{figure}[h]
    \centering
    \includegraphics[width=1.0\linewidth]{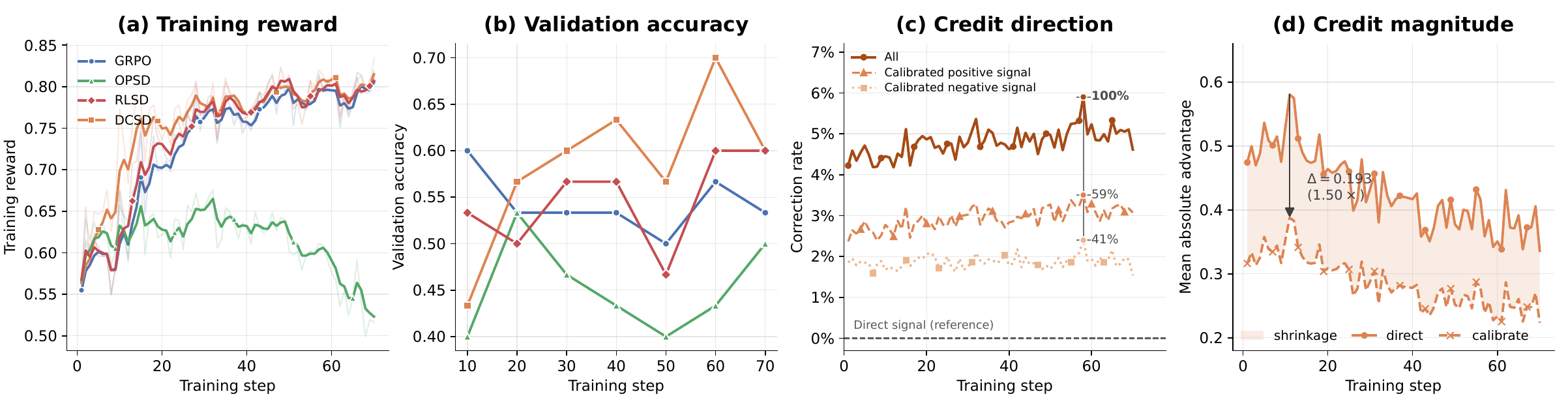}
    \caption{
\textbf{Training dynamics and empirical behavior of decoupled credit.}
(a) Training reward and (b) validation accuracy on mathematical reasoning. DCSD achieves training rewards comparable to GRPO and RLSD while attaining higher validation accuracy. (c) Direction calibration is sparse and selective, with only a small fraction of tokens receiving positive or negative overrides. (d) Magnitude calibration consistently reduces mean absolute credit, demonstrating independent control of credit direction and magnitude throughout training.}
    \label{fig3}
    
\end{figure}

\subsection{Training Dynamics and Credit Analysis}
\label{sec:dynamics}

\paragraph{Training reward and validation accuracy.}
As shown in \textbf{Figure~\ref{fig3}a-b}, DCSD, GRPO, and RLSD achieve similar training rewards throughout training but exhibit different validation performance. This gap suggests that comparable outcome rewards can lead to different generalization depending on how credit is assigned within each trajectory. DCSD concentrates updates on locally supported and informative reasoning steps, rather than propagating a common outcome direction across the entire response. This more selective credit assignment reduces reinforcement of redundant or weakly relevant steps and is consistent with the stronger validation performance of DCSD. In contrast, the later decline of OPSD shows that denser teacher supervision alone does not necessarily translate into sustained generalization when its local credit remains insufficiently calibrated.

\paragraph{Credit direction correction and magnitude control.}
As supported by \textbf{Figure~\ref{fig3}c}, direction correction occurs for about 6\% of tokens, covering both positive corrections in unsuccessful trajectories and negative corrections in successful ones. This shows that DCSD preserves most trajectory-level supervision while selectively correcting the direction when sufficient local evidence indicates otherwise. Meanwhile, in \textbf{Figure~\ref{fig3}d}, credit-magnitude attenuation remains stable throughout training, suggesting that DCSD consistently differentiates the contribution strength of individual reasoning steps. Together, these results demonstrate that DCSD independently controls \emph{which direction} a step should be updated and \emph{how much} credit it should receive.

In addition, we provide representative examples as case studies in \textit{Appendix~\ref{casestudy}}, where direct OPSD assigns the wrong credit direction to individual reasoning steps, including rewarding incorrect steps and penalizing correct ones. DCSD selectively reverses these signals, providing qualitative evidence for more reliable local credit assignment.

\subsection{Teacher Signal Deviation and Calibration Effects}
\label{sec:diagnostics}

We compare the trained models with a condition-matched Qwen3-32B reference on fixed student responses in \textbf{Table~\ref{tab:reference-mae-top1-metric-groups}}. We treat condition-matched Qwen3-32B as an operational oracle. TDR measures top-1 disagreement, while MAE measures sampled-token log-probability discrepancy. Lower values indicate closer oracle-reference alignment across (C0--C4), indicating closer agreement with the stronger-model reference in both local token decisions and sampled-token preference strengths.

\paragraph{Teacher judgment error.}
DCSD consistently achieves the lowest TDR across all privileged-information conditions, indicating fewer local top-1 disagreements with the reference. Notably, its advantage already appears under C0, where no privileged answer information is provided, showing that the improvement cannot be attributed solely to additional answer information. The advantage is maintained under correct, incorrect, equivalent, and worked-solution information (C1–C4), suggesting that DCSD remains less sensitive to changes in the privileged context. Under the proxy-oracle assumption, these results indicate lower judgment error across information conditions.

\begin{table*}[th]
\centering
\footnotesize
\setlength{\tabcolsep}{7.5pt}
\caption{
Token-level teacher-signal deviation from the condition-matched Qwen3-32B reference.
Avg.\ averages C0--C4. Lower is better; best results are \textbf{bold}.
}
\label{tab:reference-mae-top1-metric-groups}

\begin{tabular}{llccccc|c}
\toprule
Deviation & Method & C0 & C1 & C2 & C3 & C4 & Avg. \\
\midrule

\multirow{3}{*}{\shortstack[l]{Top-1 disagreement\\(TDR, \%)}}
& OPSD & 12.07 & 12.16 & 12.15 & 12.18 & 12.07 & 12.12 \\
& RLSD & 9.69 & 10.31 & 10.26 & 10.31 & 10.22 & 10.16 \\
& \cellcolor{blue!10} DCSD (\textit{Ours})
& \cellcolor{blue!10}\textbf{9.45}
& \cellcolor{blue!10}\textbf{9.95}
& \cellcolor{blue!10}\textbf{9.89}
& \cellcolor{blue!10}\textbf{9.94}
& \cellcolor{blue!10}\textbf{9.88}
& \cellcolor{blue!10}\textbf{9.82} \\

\midrule

\multirow{3}{*}{\shortstack[l]{Support discrepancy\\(MAE, $10^{-3}$)}}
& OPSD & 215.96 & 217.78 & 217.80 & 218.59 & 219.21 & 217.87 \\
& RLSD & 186.15 & 200.17 & 199.76 & 200.04 & 206.19 & 198.46 \\
& \cellcolor{blue!10} DCSD (\textit{Ours})
& \cellcolor{blue!10}\textbf{183.34}
& \cellcolor{blue!10}\textbf{193.95}
& \cellcolor{blue!10}\textbf{194.00}
& \cellcolor{blue!10}\textbf{193.89}
& \cellcolor{blue!10}\textbf{199.77}
& \cellcolor{blue!10}\textbf{192.99} \\

\bottomrule

\end{tabular}

\end{table*}

\paragraph{Teacher preference variation.}
DCSD also achieves the lowest MAE under every condition, indicating that its sampled-token preference strengths remain consistently closer to those of the reference. This advantage holds both on average and under C4, where all methods exhibit their largest deviation after receiving a complete worked solution. Thus, the improvement is not specific to a particular form of privileged information. Together, these results suggest that DCSD keeps its sampled-token preference strengths closer to the proxy oracle across privileged-information conditions, consistent with its design to constrain teacher influence after credit direction and magnitude are established.

\vspace{-4mm}

\subsection{Calibrated Teacher Signal Improves Reasoning Efficiency}
\label{sec:rlcsd}

Privileged self-distillation can overemphasize teacher-specific preferences, causing the student to imitate superficial response patterns rather than reinforce effective reasoning. RLCSD mitigates this issue by contrasting correct and incorrect references, improving reasoning performance while preserving longer reasoning traces. We therefore use RLCSD as a representative baseline to examine whether the more explicitly calibrated teacher signal in DCSD translates into greater reasoning efficiency. We jointly report Pass@1, Pass@4, and mean response length to measure single-attempt accuracy, multi-attempt solution coverage, and generation cost, respectively. 

Across six benchmarks, DCSD improves the unweighted mean Pass@1 and Pass@4 over RLCSD by 4.02\% and 3.74\%, respectively, while producing shorter responses on every benchmark and reducing mean response length by 6.3\% (\textbf{Table~\ref{tab:rlcsd_verisd_detailed}}). The reduction in generation length therefore does not come at the expense of reasoning performance; instead, DCSD achieves higher accuracy and broader solution coverage with fewer generated tokens. This accuracy-efficiency gain is consistent with teacher-signal calibration in DCSD: contribution-aware magnitude credit reduces reinforcement of redundant reasoning, while selective direction verification preserves locally useful signals.

\begin{table*}[th]
\centering
\caption{
Reasoning accuracy (\%) and efficiency (mean token length) across six benchmarks. Higher accuracy and shorter responses are better; best results are \textbf{bold}.
}
\label{tab:rlcsd_verisd_detailed}

\footnotesize
\setlength{\tabcolsep}{5pt}

\begin{tabular}{lccc|ccc|ccc}
\toprule
& \multicolumn{3}{c|}{AIME24}
& \multicolumn{3}{c|}{AIME25}
& \multicolumn{3}{c}{AIME26} \\
\cmidrule(lr){2-4}
\cmidrule(lr){5-7}
\cmidrule(lr){8-10}

Method
& Pass@1 & Pass@4 & Length
& Pass@1 & Pass@4 & Length
& Pass@1 & Pass@4 & Length \\
\midrule

RLCSD
& 63.33 & 73.33 & 11,450
& 50.00 & 70.00 & 12,946
& \textbf{60.00} & 70.00 & 12,465 \\

\rowcolor{blue!10}
DCSD (\textit{Ours})
& \textbf{70.00} & \textbf{76.67} & \textbf{10,627}
& \textbf{56.67} & \textbf{73.33} & \textbf{12,052}
& 56.67 & \textbf{76.67} & \textbf{11,847} \\

\bottomrule
\end{tabular}

\vspace{0.6em}

\begin{tabular}{lccc|ccc|ccc}
\toprule
& \multicolumn{3}{c|}{AMC23}
& \multicolumn{3}{c|}{MATH500}
& \multicolumn{3}{c}{HMMT} \\
\cmidrule(lr){2-4}
\cmidrule(lr){5-7}
\cmidrule(lr){8-10}

Method
& Pass@1 & Pass@4 & Length
& Pass@1 & Pass@4 & Length
& Pass@1 & Pass@4 & Length \\
\midrule

RLCSD
& 92.50 & \textbf{100.00} & 7,266
& 92.00 & 95.20 & 4,815
& 33.33 & 43.33 & 14,758 \\

\rowcolor{blue!10}
DCSD (\textit{Ours})
& \textbf{95.00} & 97.50 & \textbf{6,587}
& \textbf{93.60} & \textbf{96.80} & \textbf{4,164}
& \textbf{43.33} & \textbf{53.33} & \textbf{14,436} \\

\bottomrule
\end{tabular}

\end{table*}
\vspace{-4mm}

\section{Conclusion}

We study credit assignment in the self-distillation framework, showing that direct teacher supervision couples credit direction and magnitude, while practical teacher signals can deviate from oracle credit through judgment errors and preference variation. Our analysis therefore motivates treating which direction a step should be updated and how much it should contribute as two separate components of credit assignment. Based on this insight, we propose DCSD, which determines credit direction through belief-margin probing and contribution magnitude through marginal information gain, thereby calibrating privileged teacher supervision rather than directly inheriting its credit. Across 11 mathematical and multimodal reasoning benchmarks, DCSD improves overall student reasoning performance over outcome-based RL and existing self-distillation methods. More broadly, our work provides a new framework for self-distillation that enables more reliable learning from privileged teacher information through decoupled credit assignment. Future work could extend this principle to longer-horizon and interactive reasoning, where credit must be assigned across more complex sequences of decisions and feedback.

\bibliography{main}
\bibliographystyle{plainnat}


\clearpage
\appendix

\section{Related Work}\label{related work}

\paragraph{Reasoning Tasks from RLVR to OPSD.}
RLVR improves language-model reasoning by optimizing responses against outcome-level rewards. While such rewards provide reliable feedback on whether a final solution succeeds, they offer limited information about which intermediate reasoning steps are useful, harmful, or redundant. OPSD complements this trajectory-level supervision with dense token-level feedback: the same model acts as both student and privileged teacher under different conditioning contexts, allowing the teacher to evaluate the student's own rollouts with access to additional information~\citep{OPSD}. This provides finer-grained supervision without requiring a separate stronger teacher, but shifts the key challenge from obtaining dense supervision to determining whether the resulting teacher signal provides reliable local credit. Our work focuses on this latter problem by studying teacher supervision through two components of step-level credit: its \emph{direction} and \emph{magnitude}.

\paragraph{Improving Self-Distillation with Privileged Feedback.}
A natural way to improve self-distillation is to provide the teacher with richer information or make its supervision more stable. SDPO incorporates environmental feedback and successful responses into the self-teacher context, while mechanisms such as EMA or trust-region teacher regularization stabilize learning~\citep{OPSD,SDPO}. These methods demonstrate the value of privileged supervision. However, under the decomposition in Section~\ref{sec:reliable_credit}, richer feedback does not guarantee that the preference-marginalized teacher matches the success-conditioned policy, so \emph{judgment error} may remain. Moreover, temporal stability of the teacher does not imply stability of its preferences across different feedback realizations. Even with fixed teacher parameters, changes in the full conditioning context can alter local scores, allowing \emph{preference-dependent variation} to affect both credit direction and magnitude. DCSD therefore retains dense teacher evidence while determining step direction and relative magnitude independently, restricting the teacher to within-step allocation after step-level credit has been established.

\paragraph{Combining Self-Distillation with Outcome Feedback.}
Rather than relying entirely on the teacher signal, another line of work combines self-distillation with outcome-level supervision or selectively controls when and how strongly distillation is applied. RLSD anchors the update direction to the outcome advantage and uses positive teacher-derived weights to modulate token-level magnitude, combining the stability of outcome supervision with fine-grained teacher feedback. SRPO instead routes samples between GRPO and SDPO according to response correctness and the availability of teacher information, and further weights distillation positions according to teacher entropy~\citep{RLSD,SRPO}. These designs control either the source or the strength of supervision, but do not separately identify the local direction and contribution magnitude of each reasoning step. Outcome anchoring prevents the teacher from directly reversing the trajectory-level direction, but cannot recover a locally beneficial or harmful step whose contribution has the opposite sign. Teacher entropy, meanwhile, measures predictive concentration under a particular context rather than whether that confidence arises from correct task judgment; identical entropy can correspond to opposite token preferences. Consequently, positive confidence weights cannot correct an erroneous distillation direction, while their magnitude may still inherit both sources of teacher deviation. DCSD moves this calibration to the step level, selecting direction from local changes in answer belief and assigning relative magnitude from information gain in the student's own representations, rather than merely selecting or reweighting existing supervision.

\paragraph{Improving the Reliability of Teacher Supervision.}
A further line of work directly calibrates teacher supervision by comparing teacher signals across different privileged contexts. RLCSD contrasts teacher scores under correct and incorrect references to suppress shared stylistic shifts, whereas Purified OPSD subtracts a reference-only branch from the teacher conditioned on both the question and reference, followed by centering, soft clipping, and normalization to construct a purified distillation target~\citep{pan2026rlcsd,shen2026purified}. Both methods purify supervision through cross-context comparison, but sharing an outer template or reference content does not imply sharing the full conditioning context: replacing the reference reasoning and answer, or removing the question, changes how the teacher interprets the remaining content. Under our decomposition, cancelling preference-dependent effects requires the relative preference responses induced by the two full contexts to satisfy a matching condition; shared surface formatting alone does not enforce this property, and cross-context subtraction alone does not certify oracle-consistent local credit. Centering and normalization can remove vocabulary-wide common shifts, but cannot in general eliminate token-dependent context--reference--realization interactions. Moreover, the reference-only branch may itself contain useful task evidence. Thus, concentrating supervision on reasoning tokens does not by itself remove expression-dependent preferences over those tokens. DCSD does not require preference effects from different teacher contexts to cancel exactly; instead, it constrains any residual teacher deviation to redistribute credit only within a step, without changing the already established step direction or total magnitude.

\paragraph{Positioning of DCSD.}
Existing methods therefore improve on-policy self-distillation from complementary perspectives: enriching or stabilizing privileged supervision, anchoring it to outcome rewards, selectively routing or weighting distillation, and contrasting teacher signals across contexts. DCSD addresses a different question: rather than relying on a single teacher-derived signal to determine both aspects of local credit, we explicitly separate \emph{which direction} a reasoning step should be updated from \emph{how much} it should contribute. This distinction allows teacher evidence to remain useful for fine-grained token supervision while preventing teacher-dependent variation from freely determining step-level credit.

\section{Analysis of Direct Teacher Signal with Coupled Credit}
\label{app:section2_analysis}

We first define the
step-level reasoning process and its oracle local advantage, derive the
success-conditioned oracle signal, and separate the direct teacher signal
into judgment error and preference-dependent variation. We then show why
direct teacher supervision, positive outcome-anchored reweighting, and
contrastive outcome-anchored correction do not, by themselves, guarantee
oracle-consistent local credit. Throughout, \emph{direction} and \emph{magnitude} denote the sign and absolute value of the credit coefficient used in the policy objective.

\subsection{Step-Level Reasoning and Oracle Local Credit}
\label{app:s2_oracle_credit}

\paragraph{States, steps, and rollout probabilities.}
Fix the student continuation policy $\pi$, including its decoding and termination rules. Following Section~\ref{subsec:oracle_credit}, write the response as $\tau=(C_0,\ldots,C_{K-1})$, where $C_k=(y_{b_k},\ldots,y_{b_{k+1}-1})$ and $S_k=(x,C_{<k})$. Generating $C_k$ yields $S_{k+1}=(x,C_{\leq k})$, with $S_0=(x,\varnothing)$ and $S_K=(x,\tau)$. For normalized next-step identities, let $\pi(C\mid S)$ use a prefix-measurable step-ending convention or a block length fixed before continuation. The detector in Appendix~\ref{app:information_attribution_details} instead constructs its partition retrospectively. For detected blocks, the value-ratio identities apply pointwise at the realized prefixes under the original continuation policy. Normalized next-step distributions use the prefix-measurable step-ending convention specified above.

For a realized step $C_k$, its likelihood factorizes autoregressively as
\begin{equation}
\pi(C_k\mid S_k)
=
\prod_{t=b_k}^{b_{k+1}-1}
\pi(y_t\mid x,y_{<t}).
\label{eq:app_s2_block_probability}
\end{equation}
Any explicit step-ending symbol is included in the scored block; if the end position is externally fixed, the same boundary convention is used throughout.

\paragraph{Value and local advantage.}
Let $R(\tau)\in\{0,1\}$ be the terminal verifier reward, with no intermediate rewards or temporal discount. Define
\begin{equation}
\begin{aligned}
V^\pi(S)
&=
\mathbb E_\pi[R(\tau)\mid S]
=
P_\pi(R(\tau)=1\mid S),\\
Q^\pi(S,C)
&=
\mathbb E_\pi[R(\tau)\mid S,C]
=
P_\pi(R(\tau)=1\mid S,C).
\end{aligned}
\label{eq:app_s2_values}
\end{equation}
For a realized transition, $Q^\pi(S_k,C_k)=V^\pi(S_{k+1})$. Under the normalized next-step convention above, the law of total probability gives
\begin{equation}
V^\pi(S)
=
\sum_C \pi(C\mid S)Q^\pi(S,C).
\label{eq:app_s2_bellman}
\end{equation}
Hence, the oracle local credit is
\begin{equation}
A_k^\star
=
Q^\pi(S_k,C_k)-V^\pi(S_k)
=
V^\pi(S_{k+1})-V^\pi(S_k).
\label{eq:app_s2_advantage}
\end{equation}
A positive oracle advantage means that the subsequent prefix has a higher continuation success probability than the preceding prefix; a negative advantage means the reverse. Moreover,
$\mathbb E_{C\sim\pi(\cdot\mid S)}[Q^\pi(S,C)-V^\pi(S)]=0$.

Along a completed response, the local increments telescope:
\begin{equation}
\sum_{k=0}^{K-1}A_k^\star
=
V^\pi(S_K)-V^\pi(S_0)
=
R(\tau)-V^\pi(S_0).
\label{eq:app_s2_advantage_telescoping}
\end{equation}
Thus, a positive trajectory-level sum does not require every local contribution to be positive, nor does a negative sum require every contribution to be negative. A successful response may contain locally harmful steps, while an unsuccessful response may contain locally useful ones.

\paragraph{Outcome feedback is not a realized local advantage.}
The trajectory-level advantage $A_\tau$ is computed from the terminal reward and the baseline or group normalization of the underlying RL objective; it is not the oracle local advantage $A_k^\star$. A shared terminal outcome therefore does not identify the individual value differences in Eq.~\eqref{eq:app_s2_advantage}, which distinguishes trajectory-level outcome feedback from oracle local credit.

\subsection{Deriving the Success-Conditioned Oracle Signal}
\label{app:s2_success_conditioning}

\paragraph{Success conditioning and Bayes' rule.}
For $V^\pi(S)>0$, define the student policy conditioned on eventual success as
\begin{equation}
\pi^+(C\mid S)
:=
P_\pi(C\mid S,R(\tau)=1).
\label{eq:app_s2_success_policy}
\end{equation}
This reweights the student's own possible steps rather than introducing a separate expert policy. By Bayes' rule,
\begin{equation}
\begin{aligned}
\pi^+(C\mid S)
&=
\frac{
P_\pi(R(\tau)=1\mid S,C)\pi(C\mid S)
}{
P_\pi(R(\tau)=1\mid S)
}\\
&=
\pi(C\mid S)
\frac{Q^\pi(S,C)}{V^\pi(S)}.
\end{aligned}
\label{eq:app_s2_bayes}
\end{equation}
Equation~\eqref{eq:app_s2_bellman} ensures $\sum_C\pi^+(C\mid S)=1$. On the support of the student policy,
\begin{equation}
\frac{\pi^+(C\mid S)}{\pi(C\mid S)}
=
\frac{Q^\pi(S,C)}{V^\pi(S)}.
\label{eq:app_s2_success_ratio}
\end{equation}
For $\pi(C_k\mid S_k)>0$ and positive values at both boundaries, the oracle log-ratio is therefore
\begin{equation}
Z_k^\star
=
\log\frac{\pi^+(C_k\mid S_k)}{\pi(C_k\mid S_k)}
=
\log\frac{Q^\pi(S_k,C_k)}{V^\pi(S_k)}
=
\log\frac{V^\pi(S_{k+1})}{V^\pi(S_k)}.
\label{eq:app_s2_oracle_signal}
\end{equation}

\paragraph{Oracle-consistent direction.}
Because the logarithm is monotonic and $V^\pi(S_k)>0$, Eq.~\eqref{eq:app_s2_oracle_signal} gives
\begin{equation}
\operatorname{sign}(Z_k^\star)
=
\operatorname{sign}(A_k^\star).
\label{eq:app_s2_oracle_direction}
\end{equation}
This proves the directional consistency stated in Section~\ref{subsec:ideal_direction}. If $Q^\pi(S_k,C_k)=0<V^\pi(S_k)$, the negative-direction interpretation remains valid as $Z_k^\star\to-\infty$, whereas $V^\pi(S_k)=0$ makes success conditioning undefined. The finite-log analysis therefore assumes positive probabilities on the relevant support.

\paragraph{Distinguishing log-ratio and advantage scales.}
Exponentiating Eq.~\eqref{eq:app_s2_oracle_signal} gives
\begin{equation}
A_k^\star
=
V^\pi(S_k)\bigl(e^{Z_k^\star}-1\bigr),
\qquad
|A_k^\star|
=
V^\pi(S_k)\bigl|e^{Z_k^\star}-1\bigr|.
\label{eq:app_s2_oracle_magnitude}
\end{equation}
Hence, local advantage magnitude depends on both the preceding value and a nonlinear transformation of $Z_k^\star$. For example, transitions from $0.1$ to $0.2$ and from $0.4$ to $0.8$ both give $Z_k^\star=\log 2$, but advantages of $0.1$ and $0.4$, respectively. For small $Z_k^\star$,
$A_k^\star=V^\pi(S_k)Z_k^\star+O(V^\pi(S_k)(Z_k^\star)^2)$.

These are two credit coordinates linked by Eq.~\eqref{eq:app_s2_oracle_magnitude}. DCSD uses relative magnitude weights to allocate credit across steps.

\subsection{Judgment Error and Preference-Dependent Variation}
\label{app:teacher_decomposition}

\paragraph{Privileged content and preference realizations.}
Let $r$ denote the semantic content supplied to the teacher, such as an answer or reference solution, and let $u$ denote a preference realization, including its template, wording, formatting, or placement. For fixed $S$ and $r$, let $\mu(\cdot\mid S,r)$ be a distribution over semantically equivalent realizations. Define the preference-averaged teacher as
\begin{equation}
\overline{\pi}_T(C\mid S,r)
:=
\mathbb E_{u\sim\mu(\cdot\mid S,r)}
\left[\pi_T(C\mid S,r,u)\right].
\label{eq:app_s2_marginal_teacher}
\end{equation}
The average varies the realization $u$ at fixed semantic content $r$. Replacing a correct reference with an incorrect one defines a different privileged-content condition $r$.

\paragraph{Exact decomposition.}
Assume positive probability for the realized step under all policies appearing below. To expose the dependence on the preference realization, we write $Z_k^T(u)$ and $\varepsilon_k^{\mathrm{prefer}}(u)$ in this appendix, corresponding to $Z_k^T$ and $\varepsilon_k^{\mathrm{prefer}}$ in Section~\ref{subsec:practical_gap}, where the argument $u$ is suppressed. Define
\begin{equation}
\begin{aligned}
\varepsilon_k^{\mathrm{judge}}
&=
\log
\frac{\overline{\pi}_T(C_k\mid S_k,r)}
{\pi^+(C_k\mid S_k)},\\
\varepsilon_k^{\mathrm{prefer}}(u)
&=
\log
\frac{\pi_T(C_k\mid S_k,r,u)}
{\overline{\pi}_T(C_k\mid S_k,r)}.
\end{aligned}
\label{eq:app_s2_error_definitions}
\end{equation}
The realization law $\mu(\cdot\mid S,r)$ is fixed as part of the analytical reference. The first residual compares the realization-averaged teacher with the success-conditioned oracle; the second compares a particular realization with that average. Variance is taken over the specified law of $u$. Inserting these intermediate policies gives
\begin{equation}
\begin{aligned}
Z_k^T(u)
&=
\log
\frac{\pi_T(C_k\mid S_k,r,u)}
{\pi(C_k\mid S_k)}\\
&=
Z_k^\star
+
\varepsilon_k^{\mathrm{judge}}
+
\varepsilon_k^{\mathrm{prefer}}(u).
\end{aligned}
\label{eq:app_s2_teacher_decomposition}
\end{equation}
This is an exact identity rather than an approximation and requires no independence between the two deviations.

\paragraph{Preference-dependent variation.}
Holding the student, state, step, teacher parameters, and $r$ fixed while varying only $u$ gives
\begin{equation}
\begin{aligned}
Z_k^T(u)-Z_k^T(u')
&=
\varepsilon_k^{\mathrm{prefer}}(u)
-
\varepsilon_k^{\mathrm{prefer}}(u'),\\
\operatorname{Var}_{u\sim\mu}[Z_k^T(u)]
&=
\operatorname{Var}_{u\sim\mu}
[\varepsilon_k^{\mathrm{prefer}}(u)],
\end{aligned}
\label{eq:app_s2_preference_variance}
\end{equation}
whenever these variances are finite. Hence, $\varepsilon_k^{\mathrm{prefer}}(u)$ captures realization-dependent variation in the direct teacher signal. Its mean is characterized by
\begin{equation}
\mathbb E_\mu
\left[e^{\varepsilon_k^{\mathrm{prefer}}(u)}\right]
=
1,
\qquad
\mathbb E_\mu
\left[\varepsilon_k^{\mathrm{prefer}}(u)\right]
\leq0,
\label{eq:app_s2_preference_mean}
\end{equation}
where the first identity follows from Eq.~\eqref{eq:app_s2_marginal_teacher} and the second from Jensen's inequality. Thus, averaging probabilities is generally different from averaging log-probabilities.

\paragraph{Relation to token-level teacher evidence.}
For the realized step $C_k$, the teacher step probability factorizes as
$\pi_T(C_k\mid S_k,r,u)
=
\prod_{t=b_k}^{b_{k+1}-1}
\pi_T(y_t\mid x,r,u,y_{<t})$
under the same boundary convention used for the student policy. With the token-level discrepancy from Section~\ref{subsec:step_objective},
$\delta_t=
\operatorname{sg}[
\log\pi_T(y_t\mid x,r,u,y_{<t})
-
\log\pi(y_t\mid x,y_{<t})
]$,
the numerical step-level signal satisfies
\begin{equation}
Z_k^T(u)
=
\sum_{t=b_k}^{b_{k+1}-1}\delta_t.
\label{eq:app_s2_token_to_step}
\end{equation}
Stop-gradient preserves these numerical values. The identity connects step-level diagnosis to the sum of token-level teacher log-ratios. The realization average in Eq.~\eqref{eq:app_s2_marginal_teacher} is taken over complete step probabilities.

\subsection{Direct OPSD: Local Flexibility Without an Oracle Guarantee}
\label{app:s2_opsd}

\paragraph{Directional distortion.}
Direct teacher supervision allows local credit directions to vary within a response, but Eq.~\eqref{eq:app_s2_teacher_decomposition} shows that these directions need not match the oracle. For $Z_k^\star\neq0$, a sign reversal occurs exactly when
\begin{equation}
Z_k^\star
\left(
Z_k^\star
+
\varepsilon_k^{\mathrm{judge}}
+
\varepsilon_k^{\mathrm{prefer}}(u)
\right)
<0.
\label{eq:app_s2_opsd_reversal}
\end{equation}
A sufficient condition for preserving the oracle direction is
\begin{equation}
\left|
\varepsilon_k^{\mathrm{judge}}
+
\varepsilon_k^{\mathrm{prefer}}(u)
\right|
<
|Z_k^\star|.
\label{eq:app_s2_opsd_sufficient}
\end{equation}
Direct teacher supervision does not enforce this condition. When $Z_k^\star=0$, any nonzero combined deviation also produces nonzero supervision for an oracle-neutral step.

\paragraph{Log-ratio magnitude distortion.}
By the reverse triangle inequality,
\begin{equation}
\left|
|Z_k^T(u)|
-
|Z_k^\star|
\right|
\leq
\left|
\varepsilon_k^{\mathrm{judge}}
+
\varepsilon_k^{\mathrm{prefer}}(u)
\right|.
\label{eq:app_s2_log_magnitude_distortion}
\end{equation}
This bounds distortion in log-ratio magnitude, not oracle advantage magnitude. Even if $V^\pi(S_k)$ were known exactly, substituting $Z_k^T(u)$ for $Z_k^\star$ in the value conversion would yield
\begin{equation}
\begin{aligned}
&V^\pi(S_k)\left(e^{Z_k^T(u)}-1\right)-A_k^\star\\
&\qquad=
Q^\pi(S_k,C_k)
\left(
e^{\varepsilon_k^{\mathrm{judge}}
+\varepsilon_k^{\mathrm{prefer}}(u)}
-1
\right).
\end{aligned}
\label{eq:app_s2_value_conversion_error}
\end{equation}
This follows from Eq.~\eqref{eq:app_s2_teacher_decomposition} and $V^\pi(S_k)e^{Z_k^\star}=Q^\pi(S_k,C_k)$. The expression is an analytical comparison rather than an OPSD update rule: knowing the correct value scale alone does not remove error introduced by the teacher signal.

\paragraph{A normalized counterexample.}
Consider two possible steps $C$ and $D$, each sampled with probability $1/2$, with $Q^\pi(S,C)=0.8$ and $Q^\pi(S,D)=0.2$. Then $V^\pi(S)=0.5$ and $\pi^+(C\mid S)=0.8$. Let two equiprobable preference realizations satisfy $\pi_T(C\mid S,r,u_1)=0.3$ and $\pi_T(C\mid S,r,u_2)=0.9$, with the remaining probability assigned to $D$. The preference-averaged teacher therefore assigns $0.6$ to $C$. For realization $u_1$,
\begin{equation}
\begin{aligned}
Z^\star
&=\log(1.6)>0,
\qquad
A^\star=0.3,\\
\varepsilon^{\mathrm{judge}}
&=\log(0.75),
\qquad
\varepsilon^{\mathrm{prefer}}(u_1)=\log(0.5),\\
Z^T(u_1)
&=\log(0.6)<0.
\end{aligned}
\label{eq:app_s2_opsd_example}
\end{equation}
All distributions are normalized and the state value satisfies Eq.~\eqref{eq:app_s2_bellman}. Thus, sign reversal can occur within a valid reasoning decision process rather than arising from inconsistent probability assignments.

\subsection{RLSD: Outcome Anchoring and the No-Crossing Limitation}
\label{app:s2_rlsd}

\paragraph{Positive teacher reweighting.}
RLSD uses teacher evidence to modulate a trajectory-level advantage while preserving its direction~\citep{RLSD}. In our notation, its token coefficient can be written as
\begin{equation}
A_t^{\mathrm{RLSD}}
=
A_\tau m_t,
\qquad
m_t>0,
\label{eq:app_s2_rlsd_rule}
\end{equation}
where $m_t$ is a positive teacher-dependent multiplier. Averaging over the tokens of step $C_k$ gives
$A_k^{\mathrm{RLSD}}=A_\tau\overline m_k$ with $\overline m_k>0$, and therefore
\begin{equation}
\operatorname{sign}(A_k^{\mathrm{RLSD}})
=
\operatorname{sign}(A_\tau).
\label{eq:app_s2_rlsd_step}
\end{equation}
Teacher evidence can thus change credit magnitude but cannot reverse the trajectory-level direction.

\paragraph{No-crossing limitation.}
More generally, any outcome-anchored coefficient $A_k=A_\tau m_k$ with $m_k\geq0$ satisfies
\begin{equation}
A_\tau A_k\geq0.
\label{eq:app_s2_no_crossing}
\end{equation}
Hence, whenever $A_\tau A_k^\star<0$, an outcome-anchored update cannot recover the oracle local direction: $m_k>0$ gives the trajectory-level sign, while $m_k=0$ gives no update rather than the required opposite sign. The multiplier is also not determined by $|A_k^\star|$; for example, an oracle-neutral step with $A_k^\star=0$ may still receive nonzero credit when $A_\tau\neq0$. Thus, trajectory anchoring protects the outcome sign but does not identify either oracle local direction or local contribution magnitude.

\subsection{RLCSD: Contrastive Preference Cancellation and Its Limits}
\label{app:s2_rlcsd}

\paragraph{Contrastive teacher evidence.}
RLCSD contrasts positive and negative privileged references while retaining an outcome anchor~\citep{pan2026rlcsd}. For one reference in each branch, let
\begin{equation}
Z_k^{T,(\pm)}(u)
=
\log
\frac{\pi_T(C_k\mid S_k,r^\pm,u)}
{\pi(C_k\mid S_k)}.
\label{eq:app_s2_contrast_branches}
\end{equation}
Their contrast is
\begin{equation}
Z_k^{\mathrm{ctr}}(u)
=
Z_k^{T,(+)}(u)-Z_k^{T,(-)}(u)
=
\log
\frac{\pi_T(C_k\mid S_k,r^+,u)}
{\pi_T(C_k\mid S_k,r^-,u)}.
\label{eq:app_s2_contrast_signal}
\end{equation}
Applying the decomposition in Eq.~\eqref{eq:app_s2_teacher_decomposition} to both branches gives
\begin{equation}
\begin{aligned}
Z_k^{\mathrm{ctr}}(u)
&=
\varepsilon_k^{\mathrm{judge},(+)}
-
\varepsilon_k^{\mathrm{judge},(-)}+
\varepsilon_k^{\mathrm{prefer},(+)}(u)
-
\varepsilon_k^{\mathrm{prefer},(-)}(u),
\end{aligned}
\label{eq:app_s2_contrast_decomposition}
\end{equation}
Equation~\eqref{eq:app_s2_contrast_decomposition} expresses both branches relative to the same success-conditioned reference. The negative-branch residual includes its displacement from that reference, and the contrast retains task-relevant information through the remaining branch differences.

\paragraph{Residual preference variation.}
Let
\begin{equation}
\overline Z_k^{\mathrm{ctr}}
=
\log
\frac{\overline{\pi}_T(C_k\mid S_k,r^+)}
{\overline{\pi}_T(C_k\mid S_k,r^-)}.
\label{eq:app_s2_marginal_contrast}
\end{equation}
Then
\begin{equation}
Z_k^{\mathrm{ctr}}(u)-\overline Z_k^{\mathrm{ctr}}
=
\varepsilon_k^{\mathrm{prefer},(+)}(u)
-
\varepsilon_k^{\mathrm{prefer},(-)}(u).
\label{eq:app_s2_contrast_preference_residual}
\end{equation}
Complete cancellation therefore requires
$\varepsilon_k^{\mathrm{prefer},(+)}(u)
=
\varepsilon_k^{\mathrm{prefer},(-)}(u)$.
The branches may share the same outer realization while receiving different reference reasoning and answers. Their full conditioning contexts therefore differ, and shared wrapper text does not enforce equality of their relative preference effects. Matched components cancel; token- or step-dependent residuals need not.

\paragraph{Outcome anchoring and no crossing.}
RLCSD further masks and bounds its contrastive modulation while preserving the outcome-selected sign. Denoting its resulting token coefficient by $A_t^{\mathrm{RLCSD}}$, the construction satisfies
\begin{equation}
A_\tau A_t^{\mathrm{RLCSD}}\geq0.
\label{eq:app_s2_rlcsd_no_crossing}
\end{equation}
Consequently, any nonnegative aggregation of token coefficients within a step remains in the same outcome-selected half-line. When $A_\tau A_k^\star<0$, the required oracle local direction therefore cannot be recovered even if contrastive preference cancellation were exact.

Thus, contrastive teacher evidence can suppress shared preference effects, but neither guarantees their complete removal nor resolves the local-direction limitation imposed by outcome anchoring. Its contrastive scale is not, by construction, the absolute oracle advantage scale.

\subsection{Implications for Decoupled Credit for Teacher Calibration}
\label{app:s2_implications}

The preceding results identify complementary limitations rather than a general failure of self-distillation. Direct OPSD preserves local flexibility but can inherit both judgment error and preference-dependent variation. RLSD protects the trajectory-level direction but cannot recover an oracle local direction that disagrees with it. RLCSD can suppress shared preference effects, yet residual variation may remain, and its outcome-preserving modulation retains the no-crossing limitation.

These observations motivate the construction in Section~\ref{sec:method}: DCSD selects $\sigma_k$ and relative magnitude $\alpha_k$ separately from privileged teacher scores, then uses teacher evidence only for within-step allocation $q_{k,t}$. The shared response scale is $|A_\tau|$. Its token-level credit satisfies
\begin{equation}
A_t^{\mathrm{tok}}
=
\sigma_k\kappa_\tau|A_\tau|\alpha_k q_{k,t},
\qquad
\sum_{t\in C_k}A_t^{\mathrm{tok}}
=
\sigma_k\kappa_\tau|A_\tau|\alpha_k.
\label{eq:app_s2_dcsd_connection}
\end{equation}
Since $q_{k,t}>0$ and $\sum_{t\in C_k}q_{k,t}=1$, teacher variation can redistribute credit within a step but cannot alter its established direction or total magnitude. The direction certificate, evidence-based magnitude bound, and step-credit preservation property are established separately in Appendices~\ref{app:proof_direction}, \ref{app:proof_information_magnitude}, and~\ref{app:proof_teacher_isolation}, respectively. Appendix~\ref{app:dcsd_details} provides the corresponding DCSD constructions.

\section{Details of DCSD Workflow}
\label{app:dcsd_details}

This appendix provides the detailed constructions for marginal information gain, belief-margin probing, and hierarchical step--token credit allocation in DCSD. The student rollout and its trajectory-level advantage $A_\tau$ are fixed before these local credit signals are constructed.

\subsection{Detailed: From Value Estimation to Information-Attributed Credit}
\label{app:information_attribution_details}

This subsection details the representation construction underlying Section~\ref{subsec:information_gain_weighting}. After a response is generated, DCSD extracts fixed student representations and groups them into reasoning-step features used for sequential information attribution.

\paragraph{Hidden representations.}
For a generated response $\tau=(y_1,\ldots,y_T)$, let $\mathbf h_t\in\mathbb R^d$ denote the hidden representation of token $y_t$ extracted from the same fixed student layer and preprocessing procedure throughout the response. These representations are computed from the realized student trajectory and remain fixed during subsequent step induction and credit construction. Temporary transformations used by the boundary detector, such as local mean-centering, do not redefine the representations used for information attribution.

\paragraph{Step induction from local representation geometry.}
The reasoning-step partition is inferred after generation and does not alter the autoregressive rollout. For eligible positions $t$ on the stride-8 grid specified in Appendix~\ref{app:training}, the boundary between $y_{t-1}$ and $y_t$ is treated as a candidate cut. We compare adjacent windows of size $w$,
\begin{equation}
W_t^-=\{\mathbf h_{t-w},\ldots,\mathbf h_{t-1}\},
\qquad
W_t^+=\{\mathbf h_t,\ldots,\mathbf h_{t+w-1}\},
\label{eq:bd_windows}
\end{equation}
using only positions for which both windows lie fully within the response.

Following the dimension--volume view of reasoning representations~\citep{ma2026reasoningmerge}, each window is characterized by its effective dimension and local spectral volume. After subtracting the window mean, let $\{\lambda_i(W)\}$ denote the non-negative eigenvalues of the resulting Gram matrix. We define
\begin{equation}
\mathcal D(W)
=
\frac{(\sum_i\lambda_i(W))^2}
{\sum_i\lambda_i^2(W)+\epsilon},
\qquad
\mathcal V(W)
=
\frac12\sum_i
\log\!\left(
1+\frac{d}{|W|}\lambda_i(W)
\right),
\label{eq:bd_window_statistics}
\end{equation}
where $\epsilon>0$ is a numerical stabilizer and $|W|$ is the window size. The effective dimension measures how broadly variation is distributed across representation directions, while $\mathcal V(W)$ measures their regularized spectral spread. Because mean-centering removes the window mean, we additionally retain its original mean direction.

To combine these local changes, define $\Psi(W)=\mathcal V(W)-\eta_\tau\mathcal D(W)$ and construct
\begin{equation}
\boldsymbol\phi_t
=
\begin{bmatrix}
|\Psi(W_t^+)-\Psi(W_t^-)|\\
[\mathcal V(W_t^-)-\mathcal V(W_t^+)]_+\\
[\mathcal D(W_t^+)-\mathcal D(W_t^-)]_+\\
1-\cos(\boldsymbol\mu_t^-,\boldsymbol\mu_t^+)
\end{bmatrix},
\qquad
\chi_t
=
\boldsymbol\omega^\top
\operatorname{Std}_\tau(\boldsymbol\phi_t),
\label{eq:bd_transition_score}
\end{equation}
where $\boldsymbol\mu_t^-$ and $\boldsymbol\mu_t^+$ are the window means before centering, $[z]_+=\max(z,0)$, $\operatorname{Std}_\tau$ standardizes each component over eligible positions within the response, and $\boldsymbol\omega\in\mathbb R^4$ contains the detector weights.

Following the general principle of change-point detection~\citep{liu2013changepoint}, we retain high-scoring local maxima of $\chi_t$ while suppressing nearby lower-scoring candidates under a minimum-separation rule. Sorting the retained positions yields the interior boundaries $b_1,\ldots,b_{K-1}$ and hence the reasoning steps $C_0,\ldots,C_{K-1}$. The detected representation transitions define the reasoning-step units used for credit attribution.

\paragraph{Feature groups for sequential information attribution.}
Once the boundaries are fixed, each reasoning step defines the hidden-state feature group
$\mathcal H_k=\{\mathbf h_t\}_{t=b_k}^{b_{k+1}-1}$, with $\mathcal H_{<k}$ and $\mathcal H_{\leq k}$ denoting the preceding and accumulated feature groups, respectively. Across the problem and rollout distribution, these constructions induce the random feature blocks used in Theorem~\ref{thm:sequential_information_attribution}, with $X$ and $Y$ denoting the corresponding input and response random variables.

For the information identity, the fixed extraction and grouping rule induces random feature blocks across responses, as specified in Appendix~\ref{app:proof_information_attribution}. For the implemented geometry, the generated response and its realized partition are held fixed. Theorem~\ref{thm:sequential_information_attribution} then attributes to step $C_k$ the conditional response information $\mathcal I_k=I(Y;\mathcal H_k\mid X,\mathcal H_{<k})$. The proof of its marginal and telescoping properties is provided in Appendix~\ref{app:proof_information_attribution}. The computable marginal-information construction used to obtain relative credit magnitudes is developed separately in Appendix~\ref{app:magnitude_details}.

\subsection{Detailed: Belief-Margin Probe Design for Decoupled Credit Direction}
\label{app:belief_probe_details}

This subsection details the belief-margin probe introduced in Section~\ref{subsec:belief_probe}. The probe selects local direction using the student's answer scores and the known correct answer. It does not sample additional reasoning trajectories or consult privileged-context teacher probabilities to select the sign; insufficient local evidence triggers the trajectory-level fallback $\operatorname{sign}(A_\tau)$.

\paragraph{Boundary readout and answer serialization.}
At each state $S_k$, we append the same fixed answer-readout suffix $\mathcal P_{\mathrm{ans}}$ and define the readout context
\begin{equation}
\mathcal B_k=(S_k,\mathcal P_{\mathrm{ans}}).
\label{eq:bd_readout_context}
\end{equation}
The suffix requests an immediate final answer and is held fixed across all boundaries. The correct answer is not included in $\mathcal B_k$; it enters only as one of the candidate answers subsequently scored under this context.

For a semantic candidate answer $a$, let $\operatorname{ser}(a)$ be its canonical textual form and $d_{\mathrm{ans}}$ the fixed answer terminator. Its serialized token sequence is
\begin{equation}
\mathbf z(a)
=
\operatorname{Tok}\!\left(\operatorname{ser}(a)\Vert d_{\mathrm{ans}}\right)
=
(z_1,\ldots,z_{m(a)}).
\label{eq:bd_answer_serialization}
\end{equation}
The same canonicalization, serialization, and termination convention is used at every boundary. Including the terminator in the scored sequence distinguishes a complete answer from a prefix that could continue into another answer.

\paragraph{Candidate field construction.}
Candidate discovery is separated from candidate scoring. At each boundary, the next-token distribution under $\mathcal B_k$ is used only to propose likely answer heads. We retain the top-$K_d$ proposals, convert admissible proposals into complete valid answers, and denote the resulting set by $\operatorname{Discover}(S_k)$. These proposal probabilities are not used as belief scores.

Let $a^+$ denote the correct answer, $a_\tau$ the rollout answer, and $\mathcal A_\tau^{\mathrm{exp}}$ the valid answers explicitly expressed in the response. The response-level candidate field is
\begin{equation}
\mathcal A_\tau
=
\operatorname{Canon}\!\left(
\{a^+,a_\tau\}
\cup
\mathcal A_\tau^{\mathrm{exp}}
\cup
\bigcup_{k=0}^{K}\operatorname{Discover}(S_k)
\right).
\label{eq:bd_candidate_field}
\end{equation}
After construction, $\mathcal A_\tau$ is frozen and the same candidate field is scored at every boundary. A candidate discovered at a later boundary may therefore be evaluated retrospectively at an earlier boundary, but only under the earlier context $\mathcal B_k$; no later reasoning tokens are added to that context. This prevents changes in candidate membership from directly altering the belief margin. If no valid competitor to $a^+$ remains after canonicalization, the local probe is treated as inconclusive and DCSD uses the trajectory-level fallback.

\paragraph{Complete-answer scoring and belief margin.}
For each $a\in\mathcal A_\tau$, we evaluate the teacher-forced log-likelihood of its complete serialized answer:
\begin{equation}
\ell_k(a)
=
\sum_{i=1}^{m(a)}
\log
\pi\!\left(
z_i
\mid
\mathcal B_k,z_{<i}
\right).
\label{eq:bd_full_answer_score}
\end{equation}
All tokens in $\mathbf z(a)$ are prescribed by the candidate; teacher forcing supplies the preceding candidate tokens as context but does not set their probabilities to one. Thus, $\ell_k(a)$ is the likelihood of the complete answer rather than only its first token.

Using the same frozen candidate field, the correct-answer belief margin is
\begin{equation}
M_k
=
\ell_k(a^+)
-
\log\!\left(
\sum_{a\in\mathcal A_\tau\setminus\{a^+\}}
\exp(\ell_k(a))
\right).
\label{eq:bd_belief_margin}
\end{equation}
The first term measures support for the correct answer, while the second aggregates support for all competitors. Relative belief is used because an increase in $\ell_k(a^+)$ alone need not indicate improvement if competing answers increase by more.

For step $C_k$, the local belief change is
\begin{equation}
\Delta M_k=M_{k+1}-M_k.
\label{eq:bd_margin_change}
\end{equation}
Because $\mathcal B_k$ and $\mathcal B_{k+1}$ use the same readout suffix, serialization convention, and candidate field, $\Delta M_k$ isolates the change in the fixed readout associated with adding reasoning step $C_k$.

\paragraph{Direction selection.}
We use signed margin thresholds to select local directions. Theorem~\ref{thm:forward_readout_reliability} characterizes the error-tolerance conditions under which a selected direction is oracle-consistent:
\begin{equation}
\sigma_k
=
\begin{cases}
\operatorname{sign}(\Delta M_k),
& \Delta M_k\geq\tau_M^+
\ \text{or}\
\Delta M_k\leq\tau_M^-,\\
\operatorname{sign}(A_\tau),
& \text{otherwise},
\end{cases}
\label{eq:bd_direction_selection}
\end{equation}
where $\tau_M^-<0<\tau_M^+$. If no valid competitor is available, the fallback branch is used directly. 

For $\Xi_k<\min\{\tau_M^+,-\tau_M^-\}$ under the conditions of Theorem~\ref{thm:forward_readout_reliability}, each probe-selected direction satisfies $\sigma_k=\operatorname{sign}(A_k^\star)=\operatorname{sign}(Z_k^\star)$. Appendix~\ref{app:proof_direction} derives $\Xi_k$ and proves this threshold guarantee.

\paragraph{Verifier-anchored probing and empirical threshold selection.}
The belief-margin probe is anchored to the terminal verification target through the known correct answer $a^+$, which is used as a scoring target rather than inserted into the readout context.
Its local evidence is computed from the student's complete-answer likelihoods, not from a step-level verifier.
We select $\tau_M^+$ and $\tau_M^-$ empirically, with the task-specific values reported in Table~4, to control how large a readout-margin change is required before replacing the trajectory-level direction in Eq.~(49).
Theorem~2 provides a conditional guarantee: a probe-selected step is oracle-certified only when its readout-and-coverage error also satisfies $\Xi_k < \min\{\tau_M^+,-\tau_M^-\}$.
Empirical threshold selection does not by itself establish this analytical condition for every selected step, and $\Xi_k$ is not directly evaluated by the online probe.
Accordingly, ``certified'' refers to probe-selected steps satisfying the theorem's error-budget condition, rather than to all empirical overrides.
Unselected or inconclusive steps retain $\operatorname{sign}(A_\tau)$, while the step-credit preservation property in Theorem~4 holds for any fixed selected direction, independently of whether oracle certification is available.

\paragraph{Implementation.}
Complete-answer scoring can be cached and batched without changing its probabilistic definition. The model state after $\mathcal B_k$ may be reused across candidates, provided candidate sequences remain causally isolated so that each candidate attends only to $\mathcal B_k$ and its own preceding answer tokens. Such caching and batching are implementation optimizations rather than alternative scoring rules.

The probe determines $\sigma_k$, information gain determines $\alpha_k$, and privileged teacher evidence is introduced after both quantities are fixed.

\subsection{Detailed: Information-Gain Principle for Decoupled Credit Magnitude}
\label{app:magnitude_details}

This subsection details the information-gain construction introduced in Section~\ref{subsec:magnitude_control}. Given the fixed reasoning-step partition and hidden-state feature groups from Section~\ref{subsec:information_gain_weighting}, DCSD uses marginal representation information to determine the relative magnitude weight $\alpha_k$ of each step. 

\paragraph{Information-volume construction.}
Throughout this subsection, representation collections are indexed by token occurrence, so identical vectors at different positions remain distinct observations. When $\mathcal H_{<k}$ or $\mathcal H_{\leq k}$ is used as an argument of $F(\cdot)$, it denotes the indexed collection obtained by accumulating the corresponding feature groups.

For any finite representation collection $\mathcal S$, define
\begin{equation}
\mathbf B_{\mathcal S}
=
\mathbf I_d
+
\beta
\sum_{\mathbf h\in\mathcal S}
\mathbf h\mathbf h^\top,
\qquad
F(\mathcal S)
=
\frac12\log\det\mathbf B_{\mathcal S},
\label{eq:bd_volume_matrix}
\end{equation}
where $\mathbf I_d$ is the $d$-dimensional identity matrix and $\beta>0$ is fixed. If $\xi_1(\mathcal S),\ldots,\xi_d(\mathcal S)$ are the eigenvalues of $\sum_{\mathbf h\in\mathcal S}\mathbf h\mathbf h^\top$, then
\begin{equation}
F(\mathcal S)
=
\frac12
\sum_{j=1}^{d}
\log\!\left(
1+\beta\xi_j(\mathcal S)
\right).
\label{eq:bd_volume_spectrum}
\end{equation}
Thus, $F(\mathcal S)$ is a regularized log-volume of the accumulated uncentered second-moment matrix. It depends on representation energy across directions, including any shared mean component, while the identity term keeps $\mathbf B_{\mathcal S}$ positive definite.

\paragraph{Marginal information gain.}
For reasoning step $C_k$, the marginal gain beyond the preceding representation history is
\begin{equation}
\Delta F_k
=
F(\mathcal H_{\leq k})
-
F(\mathcal H_{<k}),
\label{eq:bd_step_gain}
\end{equation}
matching Eq.~\eqref{eq:conditional_information_gain} in the main text. More generally, for an incoming representation block $\mathcal C$ and preceding collection $\mathcal S$, define
\begin{equation}
\Delta F(\mathcal C\mid\mathcal S)
=
F(\mathcal S\cup\mathcal C)
-
F(\mathcal S).
\label{eq:bd_general_gain}
\end{equation}
Hence, $\Delta F_k=\Delta F(\mathcal H_k\mid\mathcal H_{<k})$.

For a single additional representation $\mathbf h$, the matrix determinant lemma gives
\begin{equation}
\Delta F(\mathbf h\mid\mathcal S)
=
\frac12
\log\!\left(
1+\beta
\mathbf h^\top
\mathbf B_{\mathcal S}^{-1}
\mathbf h
\right).
\label{eq:bd_single_gain}
\end{equation}
The history-adjusted term
$\mathbf h^\top\mathbf B_{\mathcal S}^{-1}\mathbf h$
is smaller when the incoming representation lies mainly in directions already well represented by the preceding context, and larger when it contributes comparatively novel variation.

Accordingly, the fixed log-determinant set function satisfies diminishing returns: for $\mathcal S_1\subseteq\mathcal S_2$ and a common incoming block $\mathcal C$ disjoint from both histories,
\begin{equation}
\Delta F(\mathcal C\mid\mathcal S_1)
\geq
\Delta F(\mathcal C\mid\mathcal S_2)
\geq0.
\label{eq:bd_diminishing}
\end{equation}
Thus, representation structure already captured by earlier reasoning contributes progressively less additional gain when encountered again.

\paragraph{Relative credit magnitude.}
For $\max_j\Delta F_j>0$, DCSD converts the marginal gains into bounded relative magnitude weights using maximum normalization:
\begin{equation}
\alpha_k
=
\Delta F_k
\Big/
\max_{0\leq j<K}\Delta F_j,
\qquad
0\leq\alpha_k\leq1,
\qquad
\max_k\alpha_k=1.
\label{eq:bd_normalize_gain}
\end{equation}
Hence, $\alpha_k$ scales step magnitude relative to the largest measured information gain in the same response. Unlike a normalized share, $\sum_k\alpha_k$ is not constrained to one: the maximum-gain step retains unit relative magnitude, while the remaining steps are scaled according to their information contribution. If all gains vanish, we set $\alpha_k=0$ for every $k$, so the response receives no step credit.

\paragraph{Conditional-information interpretation.}
The log-determinant construction admits an information-theoretic interpretation under an auxiliary linear-Gaussian observation model. Let $\Theta\sim\mathcal N(0,\mathbf I_d)$ and associate each hidden representation $\mathbf h$ with
\begin{equation}
O_{\mathbf h}
=
\sqrt{\beta}\,
\mathbf h^\top\Theta
+
\epsilon_{\mathbf h},
\qquad
\epsilon_{\mathbf h}\sim\mathcal N(0,1),
\label{eq:bd_auxiliary_observation}
\end{equation}
where the observation noises are mutually independent and independent of $\Theta$. For a representation collection $\mathcal S$, let
$O_{\mathcal S}=\{O_{\mathbf h}:\mathbf h\in\mathcal S\}$.
Under this model,
\begin{equation}
F(\mathcal S)
=
I(\Theta;O_{\mathcal S}),
\qquad
\Delta F_k
=
I\!\left(
\Theta;
O_{\mathcal H_k}
\mid
O_{\mathcal H_{<k}}
\right).
\label{eq:bd_auxiliary_information}
\end{equation}
Therefore, $\Delta F_k$ measures the conditional information introduced by the current representation block beyond that already available from the preceding representations. The vectors $\mathbf h_t$ are fixed loadings in the auxiliary Gaussian-observation model.

\paragraph{Relation to value-based credit.}
The realized gains telescope to the total representation log-volume as in Eq.~\eqref{eq:app_b_volume_telescoping}. Under the evidence model in Appendix~\ref{app:proof_information_magnitude}, $\Delta F_k=I(\Theta;O_{\mathcal H_k}\mid O_{\mathcal H_{<k}})$ bounds answer information and expected squared oracle advantage. The implemented weight $\alpha_k$ gives relative credit magnitude.

\subsection{Detailed: Calibrated Teacher Supervision for Step-to-Token Credit Assignment}
\label{app:step_objective_details}

This subsection details the step-to-token credit assignment introduced in Section~\ref{subsec:step_objective}. Once the student-side direction $\sigma_k$ and relative magnitude $\alpha_k$ have been fixed, privileged teacher information is used only to redistribute the resulting step-level credit among tokens within $C_k$.

\paragraph{Teacher--student token signal.}
For each sampled response token $y_t$, define
\begin{equation}
\delta_t
=
\operatorname{sg}\!\left[
\log\pi_T(y_t\mid x,r,u,y_{<t})
-
\log\pi(y_t\mid x,y_{<t})
\right],
\label{eq:bd_teacher_signal}
\end{equation}
where $\operatorname{sg}$ denotes stop-gradient and $\pi_T$ is the actual fixed teacher snapshot used for scoring. The deviation decomposition applies to this snapshot without requiring its parameters to equal the current student parameters. The student prefix and token are fixed by the rollout, so the teacher evaluates the realized student trajectory rather than generating an alternative response. Thus, $\delta_t$ supplies token-level teacher--student log-ratio evidence for within-step allocation.

\paragraph{Direction-aware teacher modulation.}
For each token $t\in C_k$, teacher evidence is aligned with the established step direction and converted into a positive bounded weight:
\begin{equation}
w_{k,t}
=
\operatorname{clip}\!\left(
e^{\sigma_k\delta_t},
1-\epsilon_w,
1+\epsilon_w
\right),
\qquad
0\leq\epsilon_w<1.
\label{eq:bd_teacher_weight}
\end{equation}
Hence, $1-\epsilon_w\leq w_{k,t}\leq1+\epsilon_w$. When $\sigma_k=+1$, tokens more strongly supported by the teacher receive larger weights; when $\sigma_k=-1$, this ordering is reversed so that relatively disfavored tokens receive a larger share of the negative credit. In either case, the teacher changes only relative token weighting within the selected step direction.

\paragraph{Within-step normalization.}
The bounded weights are normalized independently within each reasoning step:
\begin{equation}
q_{k,t}
=
w_{k,t}\Big/\sum_{s\in C_k}w_{k,s}.
\label{eq:bd_teacher_allocation}
\end{equation}
By construction, $q_{k,t}>0$ and $\sum_{t\in C_k}q_{k,t}=1$. Therefore, teacher evidence determines only the relative allocation within $C_k$ and cannot change the total step-level credit.

The clipping range also bounds how concentrated this allocation can become. Writing $n_k=|C_k|$,
\begin{equation}
\frac{1-\epsilon_w}{n_k(1+\epsilon_w)}
\leq
q_{k,t}
\leq
\frac{1+\epsilon_w}{n_k(1-\epsilon_w)}.
\label{eq:bd_allocation_bounds}
\end{equation}
Thus, arbitrarily large teacher--student likelihood ratios cannot concentrate unbounded credit on a single token. When $\epsilon_w=0$, $q_{k,t}=1/n_k$ and the allocation is uniform; when $n_k=1$, $q_{k,t}=1$ regardless of teacher evidence.

\paragraph{Token-level credit and common reference scale.}
For any fixed positive, teacher-independent common scale $\kappa_\tau$, the reference magnitude is $\kappa_\tau|A_\tau|$; its absolute value retains the outcome dependence. In all experiments we set $\kappa_\tau=T/K$, the mean step length of the response, so that the average token credit, $|A_\tau|K^{-1}\sum_k\alpha_k$, stays on the scale of the outcome advantage. Combining direction, relative magnitude, and within-step allocation gives
\begin{equation}
A_t^{\mathrm{tok}}
=
\sigma_k
\kappa_\tau|A_\tau|
\alpha_k
q_{k,t},
\qquad
t\in C_k.
\label{eq:bd_token_credit}
\end{equation}
Because $q_{k,t}$ is positive and normalized,
\begin{equation}
\sum_{t\in C_k}
A_t^{\mathrm{tok}}
=
\sigma_k\kappa_\tau|A_\tau|\alpha_k,
\qquad
\sum_{t\in C_k}
|A_t^{\mathrm{tok}}|
=
\kappa_\tau|A_\tau|\alpha_k.
\label{eq:bd_step_credit}
\end{equation}
Whenever $\alpha_k>0$ and $A_\tau\neq0$, every nonzero token credit has sign $\sigma_k$. Thus, privileged teacher information can redistribute credit within the step but cannot alter its established direction or total magnitude.

Since $\alpha_k$ is maximum-normalized rather than sum-normalized, the total absolute response credit is
\begin{equation}
\sum_{k=0}^{K-1}
\sum_{t\in C_k}
|A_t^{\mathrm{tok}}|
=
\kappa_\tau|A_\tau|
\sum_{k=0}^{K-1}\alpha_k,
\label{eq:bd_response_absolute_credit}
\end{equation}
The response-wide absolute credit is the common reference scale $\kappa_\tau|A_\tau|$ multiplied by $\sum_k\alpha_k$, as shown above.

\paragraph{Detached policy optimization.}
All quantities used to construct $A_t^{\mathrm{tok}}$ are treated as fixed credit coefficients during the policy update. In particular, gradients do not propagate through $A_\tau$, $\alpha_k$, $\sigma_k$, $\delta_t$, $w_{k,t}$, or $q_{k,t}$.

Let $\pi_{\mathrm{old}}$ denote the rollout policy and $\pi_\theta$ the policy being optimized, with probability ratio
\begin{equation}
\varrho_t
=
\frac{
\pi_\theta(y_t\mid x,y_{<t})
}{
\pi_{\mathrm{old}}(y_t\mid x,y_{<t})
}.
\label{eq:bd_policy_ratio}
\end{equation}
The detached token credit defines the following weighted token-level clipped policy surrogate:
\begin{equation}
\mathcal L_{\mathrm{DCSD}}
=
-\mathbb E_t\!\left[
\min\!\left(
\varrho_t A_t^{\mathrm{tok}},
\operatorname{clip}(\varrho_t,1-\epsilon_{\mathrm{ppo}}^{-},1+\epsilon_{\mathrm{ppo}}^{+})
A_t^{\mathrm{tok}}
\right)
\right].
\label{eq:bd_policy_loss}
\end{equation}
Here $(\epsilon_{\mathrm{ppo}}^{-},\epsilon_{\mathrm{ppo}}^{+})=(0.2,0.28)$ are the reported lower and upper clipping parameters. 
Gradients therefore flow through the current-policy ratio $\varrho_t$, not through the credit-construction procedure. DCSD changes local credit allocation without adding a separately differentiated auxiliary objective.

\paragraph{Computation order and edge cases.}
For each student rollout, DCSD first obtains the terminal reward, trajectory-level advantage, and student representations; induces the reasoning-step partition and relative magnitude weights $\alpha_k$; computes the belief-margin directions $\sigma_k$; evaluates the privileged teacher on the realized student tokens; forms $w_{k,t}$ and $q_{k,t}$; and finally constructs $A_t^{\mathrm{tok}}$ for policy optimization.

If $A_\tau=0$, all token credits are zero. If $\alpha_k=0$, step $C_k$ receives zero credit even though $q_{k,t}$ remains well defined. If all gains vanish, $\alpha_k$ follows the convention in Appendix~\ref{app:magnitude_details}. The within-step allocation $q_{k,t}$ redistributes the established step credit among its tokens.

\section{Proofs of Theoretical Results in DCSD}
\label{app:method_proofs}

This appendix proves the four Theorems in Section~\ref{sec:method} in their order of appearance. We use the notation of the main text and Appendix~\ref{app:section2_analysis}: reasoning steps are indexed by $k\in\{0,\ldots,K-1\}$, $C_k=(y_{b_k},\ldots,y_{b_{k+1}-1})$, and $S_k=(x,C_{<k})$, where $1=b_0<\cdots<b_K=T+1$. The notation $t\in C_k$ is shorthand for the token-position condition $b_k\leq t<b_{k+1}$. All logarithms are natural, and information quantities are measured in nats.

The oracle quantities retain their original definitions: $V^\pi(S)=P_\pi(R(\tau)=1\mid S)$, $A_k^\star=V^\pi(S_{k+1})-V^\pi(S_k)$, and $Z_k^\star=\log[V^\pi(S_{k+1})/V^\pi(S_k)]$ on the positive-value domain. These quantities use the original continuation policy evaluated at each realized prefix. The normalized next-step identities additionally use the prefix-based step-ending convention specified in Appendix~\ref{app:s2_oracle_credit}.

Theorem~\ref{thm:forward_readout_reliability} requires the readout-calibration and candidate-coverage conditions summarized in its statement by a finite $\Xi_k$ and formalized in Appendix~\ref{app:proof_direction}.
\subsection{Proof of Theorem~\ref{thm:sequential_information_attribution}: Sequential Information Attribution as Step Credit}
\label{app:proof_information_attribution}

\paragraph{Random variables and conditioning.}
Let $(X,\Upsilon,\Phi_0,\ldots,\Phi_{K-1})$ follow the joint law induced by the problem distribution, the fixed student policy, and a specified feature-group construction with $K$ groups, where $\Upsilon$ is the attribution target. In information expressions, $\Phi_k$ denotes a random feature block. We write $\Phi_{<k}=(\Phi_0,\ldots,\Phi_{k-1})$, $\Phi_{\leq k}=(\Phi_0,\ldots,\Phi_k)$, and $\Phi_{0:K-1}=(\Phi_0,\ldots,\Phi_{K-1})$; the empty prefix $\Phi_{<0}$ is a constant object.

Let $H$ denote conditional Shannon entropy for a discrete target, $I$ mutual information or conditional mutual information as indicated by its arguments, and $D_{\mathrm{KL}}$ Kullback--Leibler divergence. We assume $I(\Upsilon;\Phi_{0:K-1}\mid X)<\infty$, so the information quantities below are finite and their differences are well defined. Continuous-valued targets and feature blocks are permitted; their differential entropies need not exist separately.

Two instances are used. In the \emph{response instance}, $\Upsilon=Y$ is the complete model response, and $\Phi_k=\mathcal H_k$. Here $I(Y;\mathcal H_{0:K-1}\mid X)\leq H(Y\mid X)<\infty$ for a finite vocabulary and a bounded response length, and we write $\mathcal I_k=\mathcal I_k^{Y}$. In the \emph{evidence instance}, the input $X=x$ is fixed, $\Upsilon=\Theta$ is the latent answer variable of the evidence model in Appendix~\ref{app:proof_information_magnitude}, and $\Phi_k=O_{\mathcal H_k}$. Here $I(\Theta;O_{\mathcal H_{0:K-1}})=F(\mathcal H_{0:K-1})<\infty$ by Eq.~\eqref{eq:bd_auxiliary_information}.

The fixed extraction and grouping rule defines random feature blocks over the response distribution. Data-dependent boundaries and block lengths are included in these random variables. If group counts vary, use a finite common index range $0,\ldots,K_{\max}-1$, appending a distinguished empty block after the last group. The same chain-rule proof then applies on this common range.

\paragraph{Marginal target information.}
Define
\begin{equation}
\mathcal I_k^{\Upsilon}
=
I(\Upsilon;\Phi_k\mid X,\Phi_{<k}).
\label{eq:app_b_attribution_target}
\end{equation}
The conditional-information chain rule gives
\begin{equation}
I(\Upsilon;\Phi_{\leq k}\mid X)
=
I(\Upsilon;\Phi_{<k}\mid X)
+
I(\Upsilon;\Phi_k\mid X,\Phi_{<k}).
\label{eq:app_b_information_chain}
\end{equation}
Subtracting the first term yields
\begin{equation}
\mathcal I_k^{\Upsilon}
=
I(\Upsilon;\Phi_{\leq k}\mid X)
-
I(\Upsilon;\Phi_{<k}\mid X).
\label{eq:app_b_information_difference}
\end{equation}
For a discrete target with $H(\Upsilon\mid X)<\infty$, equivalently,
\begin{equation}
\mathcal I_k^{\Upsilon}
=
H(\Upsilon\mid X,\Phi_{<k})
-
H(\Upsilon\mid X,\Phi_{\leq k}).
\label{eq:app_b_information_entropy}
\end{equation}
Thus, the contribution is the reduction in target uncertainty after the preceding feature groups have already been observed.

\paragraph{Nonnegativity and equality.}
Let $P_{\Upsilon\mid X,\Phi_{\leq k}}$ and $P_{\Upsilon\mid X,\Phi_{<k}}$ denote the corresponding conditional target distributions. Conditional mutual information admits the representation
\begin{equation}
\mathcal I_k^{\Upsilon}
=
\mathbb E_{X,\Phi_{\leq k}}
\left[
D_{\mathrm{KL}}\!\left(
P_{\Upsilon\mid X,\Phi_{\leq k}}
\;\middle\|\;
P_{\Upsilon\mid X,\Phi_{<k}}
\right)
\right]
\geq0.
\label{eq:app_b_information_kl}
\end{equation}
Equality holds precisely when these conditional target distributions coincide almost surely, equivalently $\Upsilon\perp\Phi_k\mid(X,\Phi_{<k})$, where $\perp$ denotes conditional independence.

\paragraph{Sequential completeness.}
Summing Eq.~\eqref{eq:app_b_information_difference} gives
\begin{equation}
\begin{aligned}
\sum_{k=0}^{K-1}\mathcal I_k^{\Upsilon}
&=
\sum_{k=0}^{K-1}
\left[
I(\Upsilon;\Phi_{\leq k}\mid X)
-
I(\Upsilon;\Phi_{<k}\mid X)
\right]\\
&=
I(\Upsilon;\Phi_{0:K-1}\mid X),
\end{aligned}
\label{eq:app_b_information_completeness}
\end{equation}
because the empty feature prefix carries zero information. Together with nonnegativity, this proves Theorem~\ref{thm:sequential_information_attribution}. \hfill$\square$

\paragraph{Connection to information-theoretic feature attribution.}
The output-information criterion used in information-theoretic feature attribution motivates measuring explanatory features by the information they retain about the model output~\citep{chen2018learning}. Here, $X$ is the problem context and the explanatory variables are ordered internal feature groups. Applying this criterion to successive feature prefixes gives Eq.~\eqref{eq:app_b_attribution_target}. This applies the criterion to sequential feature groups. The criterion attributes information about whichever target is chosen: the response instance measures how steps construct the response, whereas the evidence instance measures evidence about the answer-relevant latent variable and is the instance connected to value in Appendix~\ref{app:proof_information_magnitude}.

For the response instance and any normalized response predictor $g(\cdot\mid X,\mathcal H_{<k})$ with finite expected log score, the corresponding prediction interpretation is
\begin{equation}
\begin{aligned}
\mathbb E\log g(Y\mid X,\mathcal H_{<k})
={}&-H(Y\mid X,\mathcal H_{<k})\\
&-\mathbb E_{X,\mathcal H_{<k}}
D_{\mathrm{KL}}\!\left(
P_{Y\mid X,\mathcal H_{<k}}
\;\middle\|\;
g(\cdot\mid X,\mathcal H_{<k})
\right).
\end{aligned}
\label{eq:app_b_information_logscore}
\end{equation}
The best attainable expected log score is the negative conditional entropy, so its improvement after adding $\mathcal H_k$ is exactly $\mathcal I_k$. This gives the expected-log-score interpretation of sequential information attribution.

\paragraph{Refinement consistency.}
Suppose $\Phi_k$ is split into consecutive groups $(\Phi_k^{(1)},\Phi_k^{(2)})$ without adding information: the pair and the original group determine one another. The chain rule gives
\begin{equation}
\begin{aligned}
I(\Upsilon;\Phi_k\mid X,\Phi_{<k})
={}&I(\Upsilon;\Phi_k^{(1)}\mid X,\Phi_{<k})\\
&+I(\Upsilon;\Phi_k^{(2)}
\mid X,\Phi_{<k},\Phi_k^{(1)}).
\end{aligned}
\label{eq:app_b_information_refinement}
\end{equation}
Refinement redistributes the raw information contribution while preserving its total. Maximum-normalized weights use the maximum associated with the resulting partition. The fixed log-determinant geometry additionally satisfies the diminishing-returns property proved in Appendix~\ref{app:proof_information_magnitude}.

\subsection{Proof of Theorem~\ref{thm:forward_readout_reliability}: Oracle-Consistent Credit Direction}
\label{app:proof_direction}

\paragraph{Continuation reference and complete-answer readout.}
Fix the student continuation policy $\pi$, including its decoding and termination rules, and a boundary pair $(S_k,S_{k+1})$. For $j\in\{k,k+1\}$, let $p_j^\pi(a)$ denote the distribution of canonical final answers obtained by freely continuing from $S_j$. Failed or unparsable completions are retained as incorrect outcomes rather than removed by renormalization. Assume one canonical correct answer $a^+$ whose probability equals verifier success, and write
\begin{equation}
\begin{aligned}
v_j&=p_j^\pi(a^+)=V^\pi(S_j)\in(0,1),\\
A_k^\star&=v_{k+1}-v_k,
\qquad
Z_k^\star=\log(v_{k+1}/v_k).
\end{aligned}
\label{eq:app_b_direction_reference}
\end{equation}
The oracle log-odds are $L_j^\star=\log[v_j/(1-v_j)]$, and their change is $D_k^\star=L_{k+1}^\star-L_k^\star$, as in the main text. Strict monotonicity of the logarithm and log-odds gives
\begin{equation}
\operatorname{sign}(D_k^\star)
=
\operatorname{sign}(A_k^\star)
=
\operatorname{sign}(Z_k^\star).
\label{eq:app_b_direction_monotonicity}
\end{equation}
The complete-answer score $\ell_j(a)$ evaluates the prescribed serialized answer after the fixed suffix $\mathcal P_{\mathrm{ans}}$ (Appendix~\ref{app:belief_probe_details}). The calibration model below relates this readout score to the original continuation probability $p_j^\pi(a)$ through a boundary offset and a bounded residual.

\paragraph{Readout calibration and candidate coverage.}
Use the same finite candidate field $\mathcal A_\tau$ at both boundaries, containing $a^+$ and at least one competitor. The readout at $S_j$ is calibrated if $\ell_j(a)$ is finite and $p_j^\pi(a)>0$ for every $a\in\mathcal A_\tau$, and there exist a real offset $\xi_j$, residuals $e_j(a)$, and a finite bound $\epsilon_j^{\mathrm{rd}}\geq0$ such that
\begin{equation}
\ell_j(a)
=
\log p_j^\pi(a)+\xi_j+e_j(a),
\qquad
|e_j(a)|\leq\epsilon_j^{\mathrm{rd}},
\qquad a\in\mathcal A_\tau.
\label{eq:app_b_readout_calibration}
\end{equation}
Each offset is common to all candidates at its boundary and need not be small. The covered incorrect-answer probability and its coverage fraction are
\begin{equation}
Q_j^-
=
\sum_{a\in\mathcal A_\tau\setminus\{a^+\}}p_j^\pi(a),
\qquad
c_j=\frac{Q_j^-}{1-v_j}\in(0,1].
\label{eq:app_b_candidate_coverage}
\end{equation}
Here $Q_j^-$ is a probability mass, distinct from the action-value function $Q^\pi$. The local analytical error bound of Theorem~\ref{thm:forward_readout_reliability} is
\begin{equation}
\Xi_k
=
2\left(\epsilon_k^{\mathrm{rd}}+\epsilon_{k+1}^{\mathrm{rd}}\right)
+
\left|\log\frac{c_{k+1}}{c_k}\right|.
\label{eq:app_b_direction_error_budget}
\end{equation}
The coverage contribution is the change in the log coverage fraction, $|\log(c_{k+1}/c_k)|$. Finite $\Xi_k$ denotes calibrated readouts at both boundaries; otherwise set $\Xi_k=\infty$. All continuation probabilities in these definitions are evaluated under the original fixed policy at the realized states, with the discovered candidate field held fixed for scoring.

\paragraph{Bounding the aggregate competitor score.}
The readout-error bound implies
\begin{equation}
e^{-\epsilon_j^{\mathrm{rd}}}Q_j^-
\leq
\sum_{a\in\mathcal A_\tau\setminus\{a^+\}}
p_j^\pi(a)e^{e_j(a)}
\leq
e^{\epsilon_j^{\mathrm{rd}}}Q_j^-.
\label{eq:app_b_competitor_sandwich}
\end{equation}
Define
\begin{equation}
d_j
=
\log\!\left(
\frac{\sum_{a\in\mathcal A_\tau\setminus\{a^+\}}
p_j^\pi(a)e^{e_j(a)}}{Q_j^-}
\right),
\qquad
|d_j|\leq\epsilon_j^{\mathrm{rd}}.
\label{eq:c2_competitor_residual}
\end{equation}
Then
\begin{equation}
\log\!\left(
\sum_{a\in\mathcal A_\tau\setminus\{a^+\}}e^{\ell_j(a)}
\right)
=
\xi_j+\log Q_j^-+d_j.
\label{eq:app_b_competitor_lse}
\end{equation}

\paragraph{Margin decomposition and local error.}
Subtracting the aggregate competitor score from $\ell_j(a^+)$ cancels $\xi_j$ and gives
\begin{equation}
\begin{aligned}
M_j
&=\log v_j-\log Q_j^-+e_j(a^+)-d_j\\
&=L_j^\star-\log c_j+\eta_j,
\end{aligned}
\label{eq:app_b_margin_decomposition}
\end{equation}
where $\eta_j=e_j(a^+)-d_j$ and $|\eta_j|\leq2\epsilon_j^{\mathrm{rd}}$. This cancellation occurs between candidate scores at a fixed boundary. Taking the difference between boundaries yields
\begin{equation}
\begin{aligned}
\Delta M_k-D_k^\star
&=-\log\frac{c_{k+1}}{c_k}+\eta_{k+1}-\eta_k,\\
|\Delta M_k-D_k^\star|
&\leq\Xi_k,
\end{aligned}
\label{eq:app_b_direction_difference_bound}
\end{equation}
which is the first claim of Theorem~\ref{thm:forward_readout_reliability}.

\paragraph{Threshold certificate.}
Use the signed thresholds $\tau_M^-<0<\tau_M^+$ from Eq.~\eqref{eq:step_direction} and assume $\Xi_k<\min\{\tau_M^+,-\tau_M^-\}$. If $\Delta M_k\geq\tau_M^+$, then
\begin{equation}
D_k^\star\geq\Delta M_k-\Xi_k
\geq\tau_M^+-\Xi_k>0.
\label{eq:c2_positive_certificate}
\end{equation}
If $\Delta M_k\leq\tau_M^-$, then
\begin{equation}
D_k^\star\leq\Delta M_k+\Xi_k
\leq\tau_M^-+\Xi_k<0.
\label{eq:c2_negative_certificate}
\end{equation}
Consequently, by Eq.~\eqref{eq:app_b_direction_monotonicity}, whenever the local branch is selected,
\begin{equation}
\sigma_k
=
\operatorname{sign}(\Delta M_k)
=
\operatorname{sign}(A_k^\star)
=
\operatorname{sign}(Z_k^\star).
\label{eq:app_b_direction_certificate}
\end{equation}
The strict error bound also covers margins exactly on either selection threshold. This proves Theorem~\ref{thm:forward_readout_reliability}. \hfill$\square$

\paragraph{Threshold interpretation.}
Equation~\eqref{eq:app_b_direction_difference_bound} places the oracle log-odds change in $[\Delta M_k-\Xi_k,\Delta M_k+\Xi_k]$. When $\Xi_k<\tau_M^+$, the interval is positive on the selected positive branch; when $\Xi_k<-\tau_M^-$, it is negative on the selected negative branch. Thus the signed thresholds specify a sufficient error-tolerance regime for oracle-consistent direction. A certified step is a probe-selected step satisfying the corresponding error-budget condition. DCSD uses this direction and constructs relative magnitude separately through $\alpha_k$.

\paragraph{Selective correction.}
For a fixed response with $A_\tau\neq0$, restrict attention to steps with $A_k^\star\neq0$. On this index set, let $\mathcal U$ contain the probe-selected steps, $\mathcal E_{\mathrm{out}}$ contain steps for which $\operatorname{sign}(A_\tau)\neq\operatorname{sign}(A_k^\star)$, and $\mathcal E_{\mathrm{DCSD}}$ contain steps for which $\sigma_k\neq\operatorname{sign}(A_k^\star)$. If every step in $\mathcal U$ satisfies the certificate, then
\begin{equation}
\mathcal E_{\mathrm{DCSD}}
=
\mathcal E_{\mathrm{out}}\setminus\mathcal U.
\label{eq:app_b_selective_correction}
\end{equation}
Selection is correct on $\mathcal U$, while unselected steps retain the outcome sign, yielding Eq.~\eqref{eq:app_b_selective_correction}.

\subsection{Proof of Theorem~\ref{thm:information_gain_magnitude}: Information Gain Bounds Credit Magnitude}
\label{app:proof_information_magnitude}
\label{app:information_value_bridge}

The information-volume construction is defined in Appendix~\ref{app:magnitude_details}. We collect its fixed-representation geometry, state the evidence model, and prove the information identity, the sharp answer-information envelope, the oracle-credit bounds, and the exact characterization of universally oracle-null steps.

\paragraph{Supporting fixed-representation geometry.}
\label{app:information_gain}
Use the matrix $\mathbf B_{\mathcal S}$ and log-volume $F(\mathcal S)$ from Eq.~\eqref{eq:bd_volume_matrix}. The scale $\beta>0$, vectors, and preprocessing remain fixed in every comparison, and collections are indexed by token occurrence. For a disjoint incoming block $\mathcal C$, let $n_{\mathcal C}=|\mathcal C|$ and let $\mathbf H_{\mathcal C}\in\mathbb R^{d\times n_{\mathcal C}}$ contain its vectors as columns. Determinant factorization and Sylvester's identity give
\begin{equation}
\begin{aligned}
\Delta F(\mathcal C\mid\mathcal S)
&=\frac12\log
\frac{\det(\mathbf B_{\mathcal S}+\beta\mathbf H_{\mathcal C}\mathbf H_{\mathcal C}^{\top})}
{\det\mathbf B_{\mathcal S}}\\
&=\frac12\log\det\!\left(
\mathbf I_{n_{\mathcal C}}
+\beta\mathbf H_{\mathcal C}^{\top}
\mathbf B_{\mathcal S}^{-1}\mathbf H_{\mathcal C}
\right)
\geq0.
\end{aligned}
\label{eq:app_b_volume_gain}
\end{equation}
The matrix added to the identity is positive semidefinite. If $\mathcal S_1\subseteq\mathcal S_2$, then $\mathbf B_{\mathcal S_2}\succeq\mathbf B_{\mathcal S_1}\succ0$ and $\mathbf B_{\mathcal S_2}^{-1}\preceq\mathbf B_{\mathcal S_1}^{-1}$, where $\succeq$ denotes the positive-semidefinite order. Congruence with $\mathbf H_{\mathcal C}$ and monotonicity of the log determinant therefore imply
\begin{equation}
0\leq\Delta F(\mathcal C\mid\mathcal S_2)
\leq\Delta F(\mathcal C\mid\mathcal S_1)
\label{eq:c3_diminishing_returns}
\end{equation}
for a common incoming block disjoint from $\mathcal S_2$. Along the realized partition,
\begin{equation}
\sum_{k=0}^{K-1}\Delta F_k
=F(\mathcal H_{0:K-1}),
\qquad F(\varnothing)=0.
\label{eq:app_b_volume_telescoping}
\end{equation}
For the fixed representation construction, accumulated evidence attenuates the marginal gain of subsequent observations in covered directions. Repeated nonzero observations can retain positive marginal gain.

\paragraph{History-relative information geometry.}
Let $\mathbf B_k=\mathbf B_{\mathcal H_{<k}}$ and $\mathbf B_{k+1}=\mathbf B_{\mathcal H_{\leq k}}$, and let $\mathbf H_k\in\mathbb R^{d\times(b_{k+1}-b_k)}$ contain the vectors of $\mathcal H_k$ as columns. Define
\begin{equation}
\begin{aligned}
\mathbf\Gamma_k
&=\mathbf B_k^{-1/2}\mathbf B_{k+1}\mathbf B_k^{-1/2}\\
&=\mathbf I_d+
\beta\mathbf B_k^{-1/2}\mathbf H_k\mathbf H_k^\top\mathbf B_k^{-1/2}
\succeq\mathbf I_d,
\end{aligned}
\label{eq:signed_readout_relative_geometry}
\end{equation}
where the inverse square root is the symmetric positive-definite one, and let $\gamma_1,\ldots,\gamma_d\geq1$ be its eigenvalues. Therefore,
\begin{equation}
\frac12\log\det\mathbf\Gamma_k
=\frac12\log\frac{\det\mathbf B_{k+1}}{\det\mathbf B_k}
=\frac12\sum_{i=1}^{d}\log\gamma_i
=\Delta F_k.
\label{eq:relative_geometry_determinant}
\end{equation}
These matrices are analytical quantities used in the proof.

\paragraph{Evidence-model conditions.}
\label{app:directional_readout_calibration}
Fix the input $x$, the scale $\beta>0$, the realized representations, and the step partition. The evidence model consists of three conditions.
(E1) \emph{Latent answer variable and token evidence.} $\Theta\sim\mathcal N(0,\mathbf I_d)$ and, for every token occurrence, $O_t=\sqrt\beta\,\mathbf h_t^\top\Theta+\epsilon_t$ with mutually independent $\epsilon_t\sim\mathcal N(0,1)$ independent of $\Theta$; these are the auxiliary observations of Eq.~\eqref{eq:bd_auxiliary_observation}, with the same representations and $\beta$ as $F$.
(E2) \emph{Answer and verifier.} The canonical final answer is $A=f(\Theta)$ for a measurable answer map $f$, and the verifier accepts exactly $a^+$, so that
\begin{equation}
R=\mathbf 1\{\Theta\in\mathcal G_x\},
\qquad
\mathcal G_x=f^{-1}(a^+).
\label{eq:app_b_success_region}
\end{equation}
(E3) \emph{Value and step evidence.} For $j\in\{k,k+1\}$,
\begin{equation}
V^\pi(S_j)=P\!\left(\Theta\in\mathcal G_x\mid O_{\mathcal H_{<j}}\right),
\label{eq:app_b_evidence_value}
\end{equation}
evaluated at the realized history evidence, and, conditional on the representations of step $k$ and on $O_{\mathcal H_{<k}}$, the step evidence $O_{\mathcal H_k}$ follows the predictive law of the model.

Condition~(E1) defines the auxiliary Gaussian model of Appendix~\ref{app:magnitude_details}; (E2)--(E3) specify its answer-relevant interpretation. Under these modeling assumptions, continuation values are posterior success probabilities and form a martingale, consistent with the same-policy Bellman identity.

\paragraph{Gaussian updating.}
Throughout the proof, information quantities and expectations are taken under the conditional law given the realized history evidence $O_{\mathcal H_{<k}}$, with the step evidence drawn from its predictive law. For a target $\Upsilon$, write
\begin{equation}
J_k(\Upsilon)
=
I\!\left(\Upsilon;O_{\mathcal H_k}\mid O_{\mathcal H_{<k}}\right).
\label{eq:app_b_step_information}
\end{equation}
Conjugacy gives the posterior law $\nu_k=\mathcal N(\mathbf m_k,\mathbf B_k^{-1})$ of $\Theta$ after $S_k$, and after step $k$,
\begin{equation}
\nu_{k+1}=\mathcal N(\mathbf m_{k+1},\mathbf B_{k+1}^{-1}),
\qquad
\mathbf m_{k+1}
=\mathbf B_{k+1}^{-1}\!\left(\mathbf B_k\mathbf m_k+\sqrt\beta\,\mathbf H_kO_{\mathcal H_k}\right).
\label{eq:app_b_posterior_update}
\end{equation}
The posterior precision $\mathbf B_{k+1}$ does not depend on the realized evidence. The predictive law of the step evidence is
\begin{equation}
O_{\mathcal H_k}\mid O_{\mathcal H_{<k}}
\sim
\mathcal N\!\left(
\sqrt\beta\,\mathbf H_k^\top\mathbf m_k,\;
\mathbf I_{n_k}+\beta\mathbf H_k^\top\mathbf B_k^{-1}\mathbf H_k
\right),
\label{eq:app_b_predictive_law}
\end{equation}
where $n_k=b_{k+1}-b_k$. By the law of total covariance, $\operatorname{Cov}(\mathbf m_{k+1}-\mathbf m_k)=\mathbf B_k^{-1}-\mathbf B_{k+1}^{-1}$. Hence the standardized belief shift $\mathbf s_k=\mathbf B_k^{1/2}(\mathbf m_{k+1}-\mathbf m_k)$ satisfies
\begin{equation}
\mathbf s_k\sim\mathcal N\!\left(0,\;\mathbf I_d-\mathbf\Gamma_k^{-1}\right),
\qquad
\mathbb E\|\mathbf s_k\|^2
=\sum_{i=1}^{d}\left(1-\gamma_i^{-1}\right).
\label{eq:app_b_belief_shift}
\end{equation}

\paragraph{Evidence identity.}
Subtracting Gaussian entropies and applying Eq.~\eqref{eq:app_b_volume_gain} with $\mathcal S=\mathcal H_{<k}$,
\begin{equation}
J_k(\Theta)
=
\frac12\log\det\!\left(\mathbf I_{n_k}+\beta\mathbf H_k^\top\mathbf B_k^{-1}\mathbf H_k\right)
=\Delta F_k
=\mathbb E\,D_{\mathrm{KL}}\!\left(\nu_{k+1}\,\|\,\nu_k\right),
\label{eq:app_b_evidence_identity}
\end{equation}
where the last equality expresses mutual information as the expected divergence of the updated posterior from the current one. The value does not depend on the realized history evidence, so $\Delta F_k$ is the attribution $\mathcal I_k^{\Theta}$ of Theorem~\ref{thm:sequential_information_attribution} in the evidence instance, and $\sum_k\Delta F_k=F(\mathcal H_{0:K-1})$ is its completeness identity. It is also the expected information gain of step $k$ regarded as an experiment on $\Theta$~\citep{lindley1956measure,chaloner1995bayesian}. For the realized update, the Gaussian divergence formula and Eq.~\eqref{eq:app_b_belief_shift} give
\begin{equation}
D_{\mathrm{KL}}\!\left(\nu_{k+1}\,\|\,\nu_k\right)
=
\Delta F_k
+\frac12\left(\|\mathbf s_k\|^2-\mathbb E\|\mathbf s_k\|^2\right).
\label{eq:app_b_realized_evidence}
\end{equation}
The fluctuation has variance $\frac12\sum_i(1-\gamma_i^{-1})^2$. Since $\log\gamma_i\geq1-\gamma_i^{-1}$, we have $\Delta F_k\geq\frac12\sum_i(1-\gamma_i^{-1})$; hence, with $d_k^{\mathrm{eff}}=\sum_i(1-\gamma_i^{-1})/\max_i(1-\gamma_i^{-1})$ for $\Delta F_k>0$, the realized divergence deviates from $\Delta F_k$ with relative standard deviation at most $\sqrt{2/d_k^{\mathrm{eff}}}$.

\paragraph{Verifier-uniform envelope.}
Since $A=f(\Theta)$, $J_k(\Theta)=J_k(\Theta,A)$, and the chain rule gives
\begin{equation}
\Delta F_k
=
J_k(A)
+
I\!\left(\Theta;O_{\mathcal H_k}\mid A,O_{\mathcal H_{<k}}\right),
\label{eq:app_b_evidence_split}
\end{equation}
with both terms nonnegative. As $R=\mathbf 1\{A=a^+\}$, data processing gives
\begin{equation}
J_k(R)\leq J_k(A)\leq\Delta F_k.
\label{eq:app_b_answer_information_bound}
\end{equation}
Mutual information equals the supremum of the mutual information over finite quantizations of either argument~\citep{cover2006elements}. Every finite quantization of $\Theta$ is a finite-valued answer map, so
\begin{equation}
\sup_f J_k\!\left(f(\Theta)\right)
=J_k(\Theta)
=\Delta F_k,
\label{eq:app_b_envelope_tightness}
\end{equation}
where the supremum is over all finite-valued measurable maps, with no fixed upper bound on the number of answer values. Thus $\Delta F_k$ bounds the step's answer information for every answer map, and no smaller verifier-independent quantity does.

\paragraph{Control of oracle credit.}
By Eq.~\eqref{eq:app_b_evidence_value}, $V^\pi(S_j)=\nu_j(\mathcal G_x)$ for $j\in\{k,k+1\}$. Because $0<V^\pi(S_k)<1$ and $\nu_k$ has a positive density, both $\mathcal G_x$ and its complement have positive Lebesgue measure; since $\nu_{k+1}$ also has a positive density, $0<V^\pi(S_{k+1})<1$ almost surely, and all logarithms below are finite. Posterior probabilities of a fixed event form a martingale, so
\begin{equation}
\mathbb E[A_k^\star]=0,
\label{eq:app_b_value_martingale}
\end{equation}
which is the same-policy Bellman identity. Pinsker's inequality for Bernoulli laws~\citep{cover2006elements} gives
\begin{equation}
(A_k^\star)^2
\leq
\frac12
D_{\mathrm{KL}}\!\left(
\operatorname{Bern}(V^\pi(S_{k+1}))
\,\big\|\,
\operatorname{Bern}(V^\pi(S_k))
\right).
\label{eq:app_b_bernoulli_pinsker}
\end{equation}
Since $V^\pi(S_{k+1})=P(R=1\mid O_{\mathcal H_{\leq k}})$, the expectation of the right-hand side is $\frac12J_k(R)$, and Eq.~\eqref{eq:app_b_answer_information_bound} yields
\begin{equation}
\mathbb E\!\left[(A_k^\star)^2\right]
\leq
\frac12J_k(R)
\leq
\frac12\Delta F_k.
\label{eq:app_b_advantage_bound}
\end{equation}
Moreover, conditional on the fixed representations and history, $V^\pi(S_{k+1})\in[0,1]$ has mean $V^\pi(S_k)$. Therefore,
\begin{equation}
\begin{aligned}
\mathbb E[(A_k^\star)^2]
&=\operatorname{Var}(V^\pi(S_{k+1}))\\
&\le V^\pi(S_k)(1-V^\pi(S_k))\le\tfrac14.
\end{aligned}
\label{eq:app_b_binary_variance_bound}
\end{equation}
Combining this with Eq.~\eqref{eq:app_b_advantage_bound} yields
$\mathbb E[(A_k^\star)^2]\le\min\{\tfrac12\Delta F_k,\tfrac14\}$.
For the log-ratio signal, Bayes' rule shows that the law of $O_{\mathcal H_k}$ given $R=1$ has density ratio $V^\pi(S_{k+1})/V^\pi(S_k)$ with respect to its unconditional law. Hence $\mathbb E[Z_k^\star\mid R=1]$ is a Kullback--Leibler divergence and is nonnegative; the same holds for $Z_k^-=\log[(1-V^\pi(S_{k+1}))/(1-V^\pi(S_k))]$ given $R=0$. Weighting the two by the prior outcome probabilities gives
\begin{equation}
V^\pi(S_k)\,\mathbb E[Z_k^\star\mid R=1]
+\bigl(1-V^\pi(S_k)\bigr)\,\mathbb E[Z_k^-\mid R=0]
=J_k(R)
\leq\Delta F_k,
\label{eq:app_b_log_ratio_bound}
\end{equation}
and therefore $\mathbb E[Z_k^\star\mid R=1]\leq\Delta F_k/V^\pi(S_k)$. Together with Eqs.~\eqref{eq:app_b_evidence_identity} and~\eqref{eq:app_b_envelope_tightness}, this establishes the information identity and oracle-credit bounds. We prove the remaining claims below.

\paragraph{Scale and redundancy.}
For $\gamma\geq1$, $\log\gamma/\gamma\leq1-\gamma^{-1}\leq\log\gamma$. Eq.~\eqref{eq:app_b_belief_shift} therefore gives
\begin{equation}
\frac{2\Delta F_k}{\lambda_{\max}(\mathbf\Gamma_k)}
\leq
\mathbb E\|\mathbf s_k\|^2
\leq
2\Delta F_k,
\label{eq:app_b_shift_sandwich}
\end{equation}
so $\Delta F_k$ also fixes the expected size of the standardized belief update, tightly when a single step changes the geometry little. By Eq.~\eqref{eq:c3_diminishing_returns}, evidence along directions already represented in the history yields a smaller envelope. If $\Delta F_k=0$, then $\mathbf H_k^\top\mathbf B_k^{-1}\mathbf H_k=0$ and hence $\mathbf H_k=0$; the step evidence is independent of $\Theta$, so $\nu_{k+1}=\nu_k$ and $A_k^\star=0$ for every answer map.

\paragraph{Oracle-null steps and the sharp information envelope.}
Fix the representations and realized history evidence. Write $A_k^\star(f)$ to expose the dependence on the answer map, and let $\nu_j=\mathcal N(\mathbf m_j,\mathbf B_j^{-1})$ be the two posteriors. If $\Delta F_k=0$, the preceding redundancy argument gives $\nu_{k+1}=\nu_k$, hence $A_k^\star(f)=0$ for every answer map.

Conversely, suppose $\Delta F_k>0$. Then $\mathbf B_{k+1}\succeq\mathbf B_k$ with unequal matrices, so there is a unit vector $\mathbf w$ such that $s_{k+1}<s_k$, where $\mu_j=\mathbf w^\top\mathbf m_j$ and $s_j^2=\mathbf w^\top\mathbf B_j^{-1}\mathbf w>0$. Consider a binary answer map accepted on the half-space $G_c=\{\theta:\mathbf w^\top\theta\ge c\}$. By (E3), $V^\pi(S_j)=\nu_j(G_c)\in(0,1)$, and
\begin{equation}
A_k^\star(c)=
\Phi_{\mathcal N}\!\left(\frac{\mu_{k+1}-c}{s_{k+1}}\right)
-\Phi_{\mathcal N}\!\left(\frac{\mu_k-c}{s_k}\right),
\label{eq:app_b_halfspace_credit}
\end{equation}
where $\Phi_{\mathcal N}$ is the standard normal distribution function. At each realized update, the normal distributions have different variances and thus different tail functions. Some finite $c$ therefore has $A_k^\star(c)\ne0$. Consequently,
\begin{equation}
\begin{aligned}
\Delta F_k=0
&\quad\Longleftrightarrow\quad \nu_{k+1}=\nu_k\\
&\quad\Longleftrightarrow\quad A_k^\star(f)=0\ \text{for every answer map }f.
\end{aligned}
\label{eq:app_b_universal_oracle_null}
\end{equation}
Furthermore, $A_k^\star(c)\to0$ as $c\to\pm\infty$, whereas it is nonzero for some finite $c$. Thus its absolute value varies with the verifier even when representations and evidence are fixed. This establishes the verifier dependence of realized oracle magnitude at fixed evidence, alongside the verifier-uniform information envelope. Finally, Eq.~\eqref{eq:app_b_envelope_tightness} identifies $\Delta F_k$ as the least verifier-independent upper bound on answer information. This completes the proof of Theorem~\ref{thm:information_gain_magnitude}. \hfill$\square$

\paragraph{From information attribution to step credit.}
The maximum-normalized weight in Eq.~\eqref{eq:bd_normalize_gain} satisfies $\alpha_k=0$ if and only if $\Delta F_k=0$, including the convention that all weights are zero when all gains vanish. Hence the relative-magnitude profile has exactly the universal oracle-null set of Eq.~\eqref{eq:app_b_universal_oracle_null}; otherwise it is proportional to the sharp answer-information envelope within the response. For $A_\tau\ne0$ and $\kappa_\tau>0$, the assigned absolute step credit $\sum_{t\in C_k}|A_t^{\mathrm{tok}}|=\kappa_\tau|A_\tau|\alpha_k$ has the same zero set. On steps certified by Theorem~\ref{thm:forward_readout_reliability}, its direction also agrees with the oracle. Theorem~\ref{thm:teacher_error_isolation} preserves both assigned quantities under teacher modulation. The oracle-credit moment bounds apply to the compared transitions under (E1)--(E3); the response-wide normalization is the separate algebraic operation defined in Eq.~\eqref{eq:bd_normalize_gain}.

\subsection{Proof of Theorem~\ref{thm:teacher_error_isolation}: Calibrated Teacher Supervision Preserves Step Credit}
\label{app:proof_teacher_isolation}

\paragraph{Fixed quantities and allocation.}
Fix the student policy, realized response, step partition, directions $\sigma_k\in\{-1,+1\}$, magnitude weights $\alpha_k\geq0$, scale $\kappa_\tau>0$, and trajectory-level advantage $A_\tau\neq0$. For the implemented maximum normalization, assume $\max_{0\leq k<K}\Delta F_k>0$. The teacher quantities may vary while these non-teacher quantities remain fixed; their dependence on the fixed input and teacher realization is suppressed as in the main text.

For each nonempty step, let $w_{k,t}$ be any finite positive weights and use the allocation in Appendix~\ref{app:step_objective_details}:
\begin{equation}
q_{k,t}
=\frac{w_{k,t}}{\sum_{s=b_k}^{b_{k+1}-1}w_{k,s}},
\qquad
A_t^{\mathrm{tok}}
=\sigma_k\kappa_\tau|A_\tau|\alpha_kq_{k,t}.
\label{eq:app_b_calibrated_credit}
\end{equation}
The clipping rule of Section~\ref{subsec:step_objective} satisfies $0<1-\epsilon_w\leq w_{k,t}\leq1+\epsilon_w$ for $0\leq\epsilon_w<1$, which is the positivity used in the main text. The preservation argument applies to every finite positive within-step weighting rule.

\paragraph{Positivity, normalization, and aggregate credit.}
The denominator in Eq.~\eqref{eq:app_b_calibrated_credit} is positive, so $q_{k,t}>0$ and $\sum_{t\in C_k}q_{k,t}=1$. Since $|\sigma_k|=1$,
\begin{equation}
\begin{aligned}
\sum_{t\in C_k}|A_t^{\mathrm{tok}}|
&=\kappa_\tau|A_\tau|\alpha_k
\sum_{t\in C_k}q_{k,t}
=\kappa_\tau|A_\tau|\alpha_k,\\
\sum_{t\in C_k}A_t^{\mathrm{tok}}
&=\sigma_k\kappa_\tau|A_\tau|\alpha_k.
\end{aligned}
\label{eq:app_b_step_credit_preservation}
\end{equation}
Both aggregates are unchanged by any variation of the positive within-step teacher weights, that is, for every teacher realization.

\paragraph{Selected-sign preservation and oracle transport.}
If $\alpha_k>0$, every factor in $A_t^{\mathrm{tok}}$ except $\sigma_k$ is strictly positive. Hence,
\begin{equation}
\operatorname{sign}(A_t^{\mathrm{tok}})=\sigma_k,
\qquad t\in C_k,\quad\alpha_k>0.
\label{eq:app_b_token_sign_preservation}
\end{equation}
If $\alpha_k=0$, all token coefficients in that step vanish and the magnitude identity remains valid. On a probe-selected step satisfying the certificate of Theorem~\ref{thm:forward_readout_reliability}, Eq.~\eqref{eq:app_b_direction_certificate} additionally gives
\begin{equation}
\operatorname{sign}(A_t^{\mathrm{tok}})
=\operatorname{sign}(A_k^\star)
=\operatorname{sign}(Z_k^\star),
\qquad t\in C_k,\quad\alpha_k>0.
\label{eq:app_b_teacher_oracle_transport}
\end{equation}
Each token coefficient inherits the oracle-consistent direction established for its parent reasoning step.

\paragraph{Evidence-scaled step magnitude.}
By Eq.~\eqref{eq:bd_normalize_gain}, the total step credit is
\begin{equation}
\kappa_\tau|A_\tau|\alpha_k
=
\frac{\kappa_\tau|A_\tau|}{\max_{0\leq j<K}\Delta F_j}\,\Delta F_k,
\label{eq:app_b_evidence_scaled_magnitude}
\end{equation}
which is proportional to $\Delta F_k$ with a factor common to all steps of the response. Under the evidence model, Theorem~\ref{thm:information_gain_magnitude} identifies $\Delta F_k$ as the exact latent-answer information gain and the sharp verifier-independent envelope for answer information, and bounds the second moment of oracle advantage. Its universal oracle-null characterization transfers to $\alpha_k$. These identities prove Theorem~\ref{thm:teacher_error_isolation}. \hfill$\square$

\paragraph{Within-step teacher modulation.}
For nonzero step credit and positions $t,s\in C_k$,
\begin{equation}
\frac{|A_t^{\mathrm{tok}}|}{|A_s^{\mathrm{tok}}|}
=\frac{q_{k,t}}{q_{k,s}}
=\frac{w_{k,t}}{w_{k,s}}.
\label{eq:app_b_teacher_within_step_ratios}
\end{equation}
Normalization preserves the ratios of the bounded direction-adjusted weights. A common positive multiplier cancels within a step; token-specific changes act through relative allocation, and clipping may produce ties.

\paragraph{Response-wide magnitude and edge cases.}
Summing the absolute step credits yields
\begin{equation}
\sum_{t=1}^{T}|A_t^{\mathrm{tok}}|
=\kappa_\tau|A_\tau|\sum_{k=0}^{K-1}\alpha_k.
\label{eq:app_b_response_magnitude}
\end{equation}
Thus, $\kappa_\tau|A_\tau|$ is the common reference scale, and total absolute credit follows Eq.~\eqref{eq:app_b_response_magnitude}. If all gains are zero, $\alpha_k$ follows Appendix~\ref{app:magnitude_details}.

\paragraph{Relation to teacher deviations.}
With non-teacher quantities fixed, teacher variation acts through within-step allocation while Eq.~\eqref{eq:app_b_step_credit_preservation} preserves the established step direction and total magnitude. The algebraic preservation holds for positive normalized weights; Theorem~\ref{thm:forward_readout_reliability} supplies oracle-consistent direction, and Theorem~\ref{thm:information_gain_magnitude} supplies the evidence-model magnitude interpretation.


\section{Experimental Details and Result Analysis Protocols}
\label{app:experimental-details}

This appendix documents the training configuration, the construction and
scoring of diagnostics, uncertainty estimation and runtime
controls, and the definitions of the training-dynamics statistics.

\subsection{Training Configuration}
\label{app:training}

Table~\ref{tab:app-training} summarizes the settings for mathematical
reasoning (Math) and vision--language reasoning (VL).

\begin{table}[!htbp]
\centering
\caption{Training and validation settings for mathematical and
vision--language reasoning.}
\label{tab:app-training}
\small
\renewcommand{\arraystretch}{1.10}
\begin{tabular}{@{}>{\raggedright\arraybackslash}p{0.43\linewidth}
>{\raggedright\arraybackslash}p{\dimexpr0.285\linewidth-2\tabcolsep\relax}
>{\raggedright\arraybackslash}p{\dimexpr0.285\linewidth-2\tabcolsep\relax}@{}}
\toprule
Setting & Math & VL \\
\midrule
Base model
& Qwen3-4B & Qwen3-VL-8B-Instruct \\

Training corpus
& DAPO-17K & MMFineReason-123K \\

Prompt batch / rollouts per prompt
& 256 / 4 & 256 / 8 \\

Prompt / response token limits
& 2,048 / 16,384 & 4,096 / 4,096 \\

Rollout temperature / top-$p$ / top-$k$
& $1.0 / 1.0$ / off & $1.0 / 1.0$ / off \\

Validation temperature / top-$p$ / top-$k$
& $0.7 / 0.95 / 20$ & $0.7 / 0.8 / 20$ \\

Optimizer / learning rate
& AdamW / $10^{-6}$ & AdamW / $10^{-6}$ \\

Weight decay / LR schedule / warm-up
& $0.01$ / constant / 0 & $0.01$ / constant / 0 \\

Actor global batch / PPO epochs
& 256 / 1 & 256 / 1 \\

PPO clip (lower / upper) / gradient norm
& $0.2 / 0.28$ / $1.0$ & $0.2 / 0.28$ / $1.0$ \\

Reference-policy KL loss
& Disabled & Disabled \\

Teacher refresh interval
& Every 20 updates & Every 10 updates \\

Teacher ratio clipping
& $[0.8,1.2]$ & $[0.8,1.2]$ \\

DCSD positive margin threshold $\tau_M^+$ (nat)
& $+5$ & $+6$ \\

DCSD negative margin threshold $\tau_M^-$ (nat)
& $-3$ & $-6$ \\

Image pixel range
& Not applicable & $262{,}144$--$1{,}048{,}576$ \\
\bottomrule
\end{tabular}
\end{table}
\vspace{-6mm}
\paragraph{Method-specific supervision.}
OPSD, RLSD, and DCSD use the gold final answer as privileged information,
without a worked solution. GRPO uses outcome-based group advantages.
Disabling the reference-policy KL loss does not disable the distillation
objective used by OPSD. RLCSD uses a binary answer-verification reward and
constructs its correct and incorrect references from verified
\emph{other rollouts of the same question}, excluding the response being
scored and filtering unusable groups. Effective updates use the resulting eligible samples.
\vspace{-4mm}
\paragraph{Step segmentation.}
The segmentation configuration uses a window size of 32 tokens, a stride
of 8 tokens, a minimum segment length of 24 tokens, a percentile setting
of 85, and a snapping radius of 8 tokens.

\subsection{Construction of the C0--C4 Diagnostics}
\label{app:diagnostics}

We freeze one Qwen3-4B thinking response for each of the 90 questions
from AIME24--26, using the mathematical decoding settings described above.
The stored responses contain 1,098,216 tokens per scoring condition.
All evaluated models, including Qwen3-32B, teacher-force the same stored
response token IDs, without regenerating or retokenizing the responses.
\vspace{-2mm}
\begin{table}[!htbp]
\centering
\caption{Privileged evidence supplied under the C0--C4 diagnostic
conditions. Here $g$ is the gold answer and $w$ is a deterministically
constructed incorrect answer.}
\label{tab:app-conditions}
\small
\renewcommand{\arraystretch}{1.10}
\begin{tabular}{@{}>{\raggedright\arraybackslash}p{0.13\linewidth}
>{\raggedright\arraybackslash}p{\dimexpr0.87\linewidth-2\tabcolsep\relax}@{}}
\toprule
Condition & Evidence \\
\midrule
C0 & Original question, without the additional user-message wrapper. \\

C1 & \texttt{The correct final answer is \{g\}.} \\

C2 & \texttt{The correct final answer is \{w\}.} \\

C3 & \texttt{The verified final result for this problem is \{g\}.} \\

C4 & A worked solution, followed by a blank line and the C1 sentence. \\
\bottomrule
\end{tabular}
\end{table}
\vspace{-6mm}

\paragraph{Incorrect-answer construction.}
For C2, the incorrect answer is generated deterministically from the gold
answer $g$ and the question identifier $\mathrm{id}$:
\begin{align}
d
&=1+\Bigl(
\operatorname{uint32be}\!\left(
\operatorname{SHA256}(\texttt{C2:}+\mathrm{id})[0{:}4]
\right)\bmod999
\Bigr),
\label{eq:app-c2-offset}\\
w&=(g+d)\bmod1000.
\label{eq:app-c2-answer}
\end{align}
Here, $+$ inside the hash input denotes string concatenation, and
$\operatorname{uint32be}$ interprets the first four hash bytes as an
unsigned big-endian integer. The offset lies in $\{1,\ldots,999\}$,
ensuring $w\ne g$ for the three-digit answer range.
\vspace{-4mm}
\paragraph{Equivalent wording and worked solutions.}
C3 changes only the wording of the correct-answer statement. C4 uses
AIME 2024-2026 worked solutions.

Let $y_{i,t}$ denote token $t$ of the stored response to question $i$,
and let $T_i$ be its length. For model $M$ and condition
$c\in\{0,\ldots,4\}$, define
\begin{align}
L_{M,c,i,t}
&=\log p_M\!\left(y_{i,t}\mid C_c(x_i),y_{i,<t}\right),
\label{eq:app-diagnostic-logprob}\\
v_{M,c,i,t}
&=\arg\max_v p_M\!\left(v\mid C_c(x_i),y_{i,<t}\right),
\label{eq:app-diagnostic-top1}
\end{align}
where $C_c(x_i)$ denotes the scoring prompt for question $i$ under condition $c$. Following the main-text convention, $M_{\mathrm{ref}}=\text{Qwen3-32B}$ is the condition-matched operational oracle for these diagnostics. Let $N=\sum_iT_i$ be the total number of scored response tokens. The two metrics are
\begin{align}
\mathrm{MAE}_{M,c}
&=\frac{1}{N}\sum_{i,t}
\left|L_{M,c,i,t}-L_{M_{\mathrm{ref}},c,i,t}\right|,
\label{eq:app-diagnostic-mae}\\
\mathrm{TDR}_{M,c}
&=\frac{100}{N}\sum_{i,t}
\mathbf{1}\!\left[
 v_{M,c,i,t}\ne v_{M_{\mathrm{ref}},c,i,t}
\right].
\label{eq:app-diagnostic-tdr}
\end{align}
\vspace{-1mm}

MAE uses the log-probability assigned to the actual response token and
is measured in nat/token; tabulated values are displayed in
$10^{-3}$ nat/token. TDR is the percentage of positions with different full-vocabulary top-1 predictions. Both metrics include all response
tokens and apply no base-model subtraction. The reported average gives
equal weight to C0--C4. Computing these metrics requires only stored
selected-token log-probabilities and top-1 token IDs; full probability
distributions need not be retained.
\vspace{-4mm}
\paragraph{Effect sizes.}
DCSD achieves lower MAE and TDR under every condition. Its
condition-averaged MAE is $0.192993$ nat/token, compared with $0.198465$
for RLSD and $0.217870$ for OPSD. The corresponding TDR values are
$9.8218\%$, $10.1566\%$, and $12.1243\%$, respectively.
For either error metric $E$, let
$\bar E_M=\frac{1}{5}\sum_{c=0}^{4}E_{M,c}$.
The relative reduction against a baseline is
\begin{equation}
\mathrm{Reduction}(E)
=100\left(1-
\frac{\bar E_{\mathrm{DCSD}}}{\bar E_{\mathrm{baseline}}}
\right)\%.
\label{eq:app-relative-error-reduction}
\end{equation}
Table~\ref{tab:app-reductions} reports these reductions and their paired
bootstrap intervals.
\vspace{-4mm}
\begin{table}[htbp]
\centering
\caption{Relative reduction in reference-model error. Brackets contain
paired-bootstrap 95\% confidence intervals, expressed in percent.}
\label{tab:app-reductions}
\small
\renewcommand{\arraystretch}{1.10}
\begin{tabular}{@{}lcc@{}}
\toprule
Baseline & MAE reduction & TDR reduction \\
\midrule
OPSD & $11.42\%\;[9.84,13.20]$ & $18.99\%\;[17.65,20.48]$ \\
RLSD & $2.76\%\;[2.51,3.02]$ & $3.30\%\;[3.01,3.58]$ \\
\bottomrule
\end{tabular}
\end{table}
\vspace{-4mm}
\paragraph{Oracle-relative diagnostics.}
Under the operational-oracle convention, TDR measures local token-decision disagreement and MAE measures sampled-token support discrepancy. Improvement in both indicates closer agreement with the oracle reference. C1/C3 preserve answer information while changing its wording; C0/C1/C2/C4 vary the supplied information content. For each condition, the evaluated model and the operational oracle receive the corresponding matched prompt.
\vspace{-4mm}
\paragraph{Scoring environment.}
C1--C4 are scored using two H100 GPUs with tensor parallelism of two
(TP=2) and BF16 precision. The software stack consists of vLLM
\texttt{0.8.5.post1}, PyTorch \texttt{2.6.0+cu124}, Transformers
\texttt{4.51.3}, and FlashAttention \texttt{2.7.4.post1}.
The Qwen3-32B revision has the prefix \texttt{9216db5781bf}.
The maximum input length is 17,461 tokens, within the configured
32,768-token context window, and no input is truncated.

\vspace{-4mm}
\subsection{Training-Dynamics Definitions}
\label{app:dynamics}

\paragraph{Training reward and validation accuracy.}
Training reward denotes answer
accuracy. Validation uses the 30 AIME25 questions, with one response
per question and the mathematical evaluation-decoding settings.

\vspace{-4mm}
\paragraph{Credit magnitude.}
From the DCSD training logs, we compare the mean absolute token advantage before ($s=\mathrm{direct}$, $A_t^{\mathrm{direct}}=A_\tau$) and after ($s=\mathrm{calibrate}$, $A_t^{\mathrm{calibrate}}=A_t^{\mathrm{tok}}$) magnitude allocation:
\begin{equation}
\mathrm{Mag}^{s}
=\frac{\sum_t m_t\left|A_t^{s}\right|}{\sum_t m_t},
\qquad s\in\{\mathrm{direct},\mathrm{calibrate}\},
\label{eq:app-sc-magnitude}
\end{equation}
where $m_t$ is the valid-response-token mask. 

The logged field \texttt{correction\_rate} is the fraction of valid response tokens whose selected direction $\sigma_k$ differs from $\operatorname{sign}(A_\tau)$, counting both positive and negative corrections.
\vspace{-4mm}
\subsection{Baseline Training Settings and Result Sources}
\label{app:baseline-training}
Table~\ref{tab:math-baseline-settings} documents our mathematical baseline runs. All four methods use Qwen3-4B in thinking mode and the same shuffled DAPO-17K data, prompt batches, and rollout budget. The prepared data contain 17,917 source records; final evaluation uses the step-70 checkpoints.
\vspace{-1mm}
\begin{table}[!htbp]
\centering
\caption{Training settings of the mathematical baselines implemented in this work. Top-$k$ ``off'' corresponds to $-1$. Training-time validation is not the final four-response evaluation.}
\label{tab:math-baseline-settings}
\small
\setlength{\tabcolsep}{4pt}
\renewcommand{\arraystretch}{1.07}
\begin{tabular}{@{}p{0.32\linewidth}*{4}{p{\dimexpr0.17\linewidth-2\tabcolsep\relax}}@{}}
\toprule
Setting & GRPO & OPSD & RLSD & RLCSD \\
\midrule
Base model / training data & \multicolumn{4}{c}{Qwen3-4B (thinking) / DAPO-17K} \\
Epochs / evaluated checkpoint & \multicolumn{4}{c}{1 / step 70} \\
Prompt batch / rollouts per prompt & \multicolumn{4}{c}{256 / 4} \\
Prompt / response limits (tokens) & \multicolumn{4}{c}{2,048 / 16,384} \\
Train temperature / top-$p$ / top-$k$ & \multicolumn{4}{c}{$1.0$ / $1.0$ / off} \\
Validation $T$ / top-$p$ / top-$k$ / $n$ & \multicolumn{4}{c}{$0.7$ / $0.95$ / $20$ / $1$} \\
Optimizer / learning rate & \multicolumn{4}{c}{AdamW / $10^{-6}$} \\
Adam betas / weight decay & \multicolumn{4}{c}{$(0.9,0.999)$ / $0.01$} \\
LR schedule / warm-up & \multicolumn{4}{c}{Constant / 0} \\
Actor batch parameter / passes & \multicolumn{4}{c}{256 / 1} \\
Gradient norm clip / seed & \multicolumn{4}{c}{$1.0$ / $1$} \\
PPO clip (lower, upper) & $(0.2,0.28)$ & -- & $(0.2,0.28)$ & $(0.2,0.28)$ \\
Reference-policy KL coefficient & $0$ & $0$ & $0$ & $0$ \\
Privileged context & None & Gold answer & Gold answer & Other rollouts \\
Teacher weights / refresh & -- & Snapshot / 20 & Snapshot / 20 & Snapshot / 20 \\
RLSD initial $\lambda$ / decay & -- & -- & $0.5$ / 60 & -- \\
RLSD ratio clipping & -- & -- & $[0.8,1.2]$ & -- \\
\bottomrule
\end{tabular}
\vspace{-6mm}
\end{table}
\vspace{-6mm}
\paragraph{Method-specific settings.}
 OPSD uses sampled-token self-distillation with coefficient $1$. RLSD uses a frozen teacher refreshed every 20 iterations, with $\lambda$ decaying from $0.5$ to zero over 60 steps and no warm-up. RLCSD uses binary answer reward and a snapshot refreshed every 20 iterations. Its archived parameters are $(\tau,\beta,\lambda,\delta,\eta)=(0.02,1,0.5,0.02,1)$, residual clipping $[-2,2]$, and $K_{\max}=4$; token-level rollout importance sampling is capped at $2$. Correct and incorrect contexts come from eligible same-question rollouts, excluding the target. Groups without usable contexts are skipped. Reference-policy KL being disabled does not disable OPSD's distillation objective. Multimodal evaluation. We use the evaluation script from the RLSD source code to ensure alignment.

\subsection{Computational Overhead}\label{computational overhead}
All mathematical experiments were trained and evaluated on two NVIDIA
H100 SXM GPUs. Table~\ref{tab:training-cost-breakdown} reports the
logged computational cost per training step. DCSD introduces additional
post-processing to extract and analyze hidden representations, induce
reasoning steps, compute marginal information gains, and perform
belief-margin probing. Consequently, its other-processing cost is
$0.3463$ GPU-hours per step. Nevertheless, generation remains the
dominant cost shared across methods, so the increase in end-to-end
training cost is considerably smaller: DCSD requires $1.6481$ GPU-hours
per step, only $23.2\%$ more than OPSD. Given the performance gains reported in
Section~\ref{sec:main_results}, this additional computation yields a favorable
performance--compute trade-off.
\begin{table}[!b]

\centering
\footnotesize
\setlength{\tabcolsep}{8pt}
\renewcommand{\arraystretch}{0.88}
\caption{Logged training cost (GPU-hours per step) for mathematical reasoning.}
\label{tab:training-cost-breakdown}
\vspace{-1.5mm}
\begin{tabular}{@{}lccccc@{}}
\toprule
Method & Generation & Old log-$p$ & Actor & Other & Total \\
\midrule
GRPO
& 0.8297 & 0.0760 & 0.3089 & 0.0008 & 1.2154 \\
OPSD
& 0.9775 & -- & 0.3603 & $<0.0001$ & 1.3379 \\
RLSD
& 0.8516 & 0.0661 & 0.3280 & 0.0027 & 1.2484 \\
RLCSD$^{*}$
& 0.8471 & 0.0823 & 0.0618 & 0.0923 & 1.0835 \\
DCSD$^{\dagger}$
& 0.8505 & 0.0748 & 0.3765 & 0.3463 & 1.6481 \\
\bottomrule
\end{tabular}
\vspace{-2mm}
\end{table}

\clearpage

\section{Case Studies}
\label{casestudy}
Illustrative cases are selected for the displayed sign contrasts or prompt responses. Reported span scores refer to the black-framed tokens; green and red indicate positive and negative signals, respectively. The captions assess local semantic contribution: correct useful steps, erroneous steps, and purely stylistic or repeated content.

\begin{figure}[!ht]
\centering
\includegraphics[width=\linewidth,height=0.62\textheight,keepaspectratio]{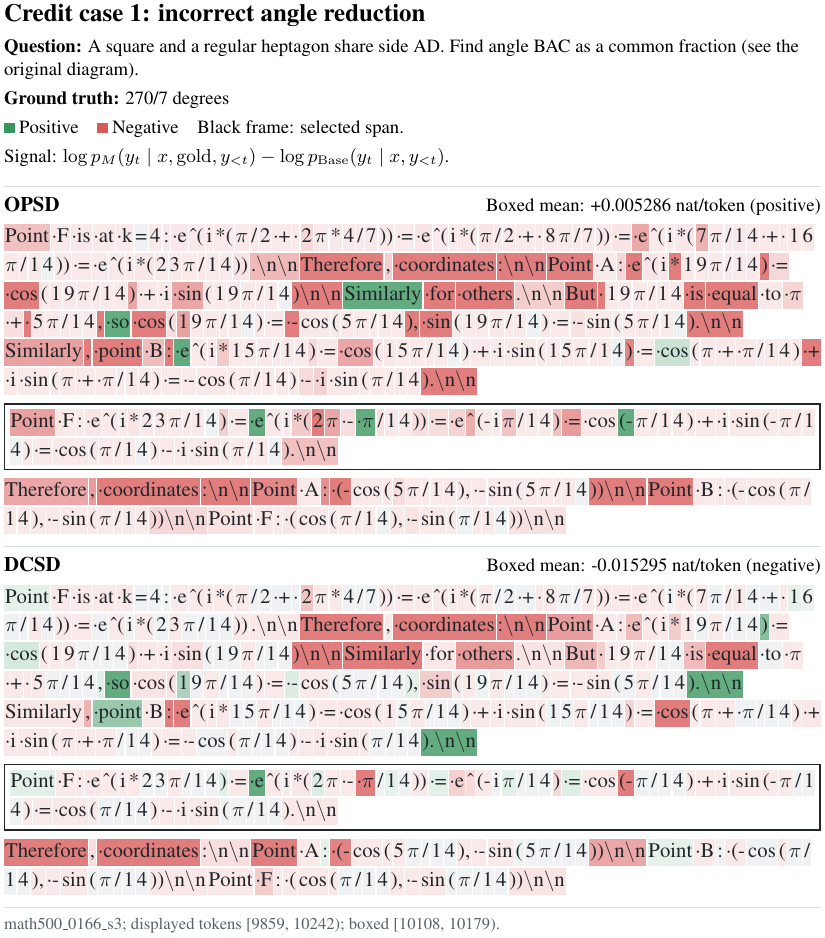}
\caption{\textbf{Incorrect angle reduction.} The boxed chain incorrectly replaces $23\pi/14$ by $2\pi-\pi/14$; the correct reduction is $2\pi-5\pi/14$. The ensuing trigonometric coordinates therefore follow an incorrect angle. This error warrants negative local credit under the semantic criterion used here. OPSD assigns a positive mean signal ($+0.005286$), whereas DCSD assigns a negative one ($-0.015295$), agreeing with the error judgment. Values are means over the same boxed tokens in nat/token.}
\label{fig:credit-case-01}
\end{figure}\clearpage

\begin{figure}[!ht]
\centering
\includegraphics[width=\linewidth,height=0.80\textheight,keepaspectratio]{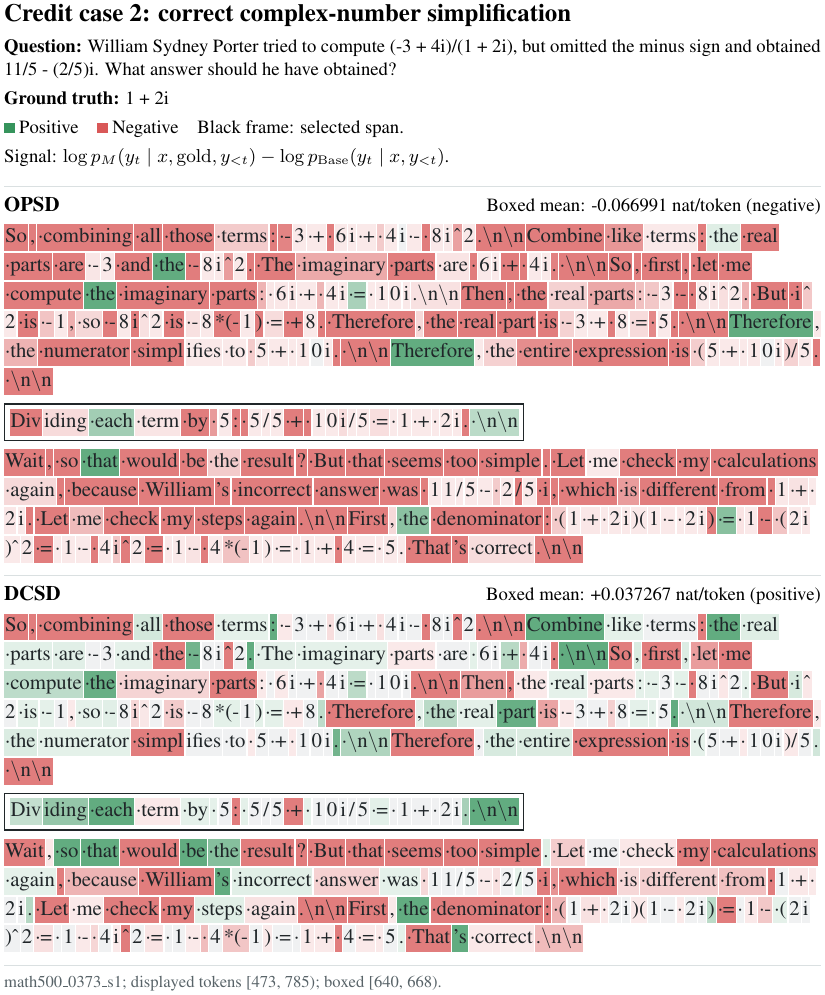}
\caption{\textbf{Correct complex-number simplification.} The boxed calculation $(5+10i)/5=1+2i$ is valid and completes the correct evaluation of $(-3+4i)/(1+2i)$. It warrants positive local credit as a useful correct step. OPSD instead assigns negative mean credit ($-0.066991$); DCSD assigns positive mean credit ($+0.037267$), consistent with this semantic judgment. Both panels score the identical frozen response with the same gold-answer information and Base reference.}
\label{fig:credit-case-02}
\end{figure}\clearpage

\begin{figure}[!ht]
\centering
\includegraphics[width=\linewidth,height=0.80\textheight,keepaspectratio]{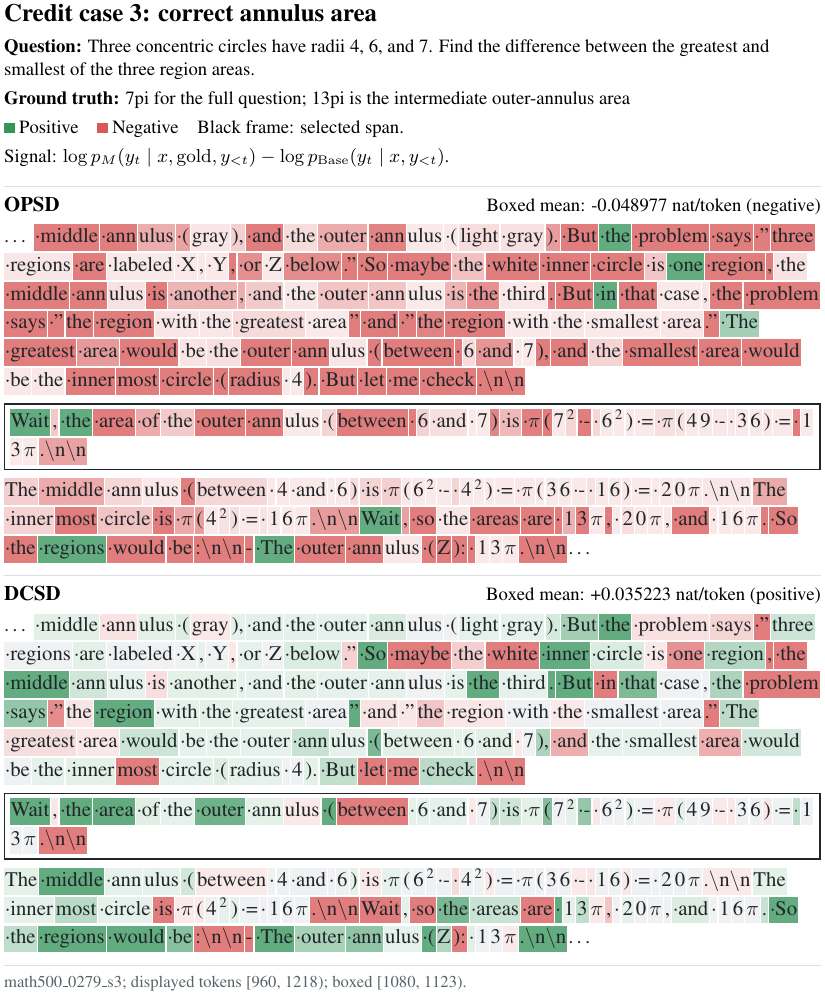}
\caption{\textbf{Correct intermediate area.} The boxed expression $\pi(7^2-6^2)=13\pi$ correctly computes the outer annulus. This is a useful intermediate quantity, not the final answer: the three areas are $16\pi$, $20\pi$, and $13\pi$, so the requested difference is $7\pi$. The boxed step warrants positive local credit. OPSD assigns $-0.048977$, while DCSD assigns $+0.035223$, matching the positive semantic judgment.}
\label{fig:credit-case-03}
\end{figure}\clearpage

\begin{figure}[!ht]
\centering
\includegraphics[width=\linewidth,height=0.80\textheight,keepaspectratio]{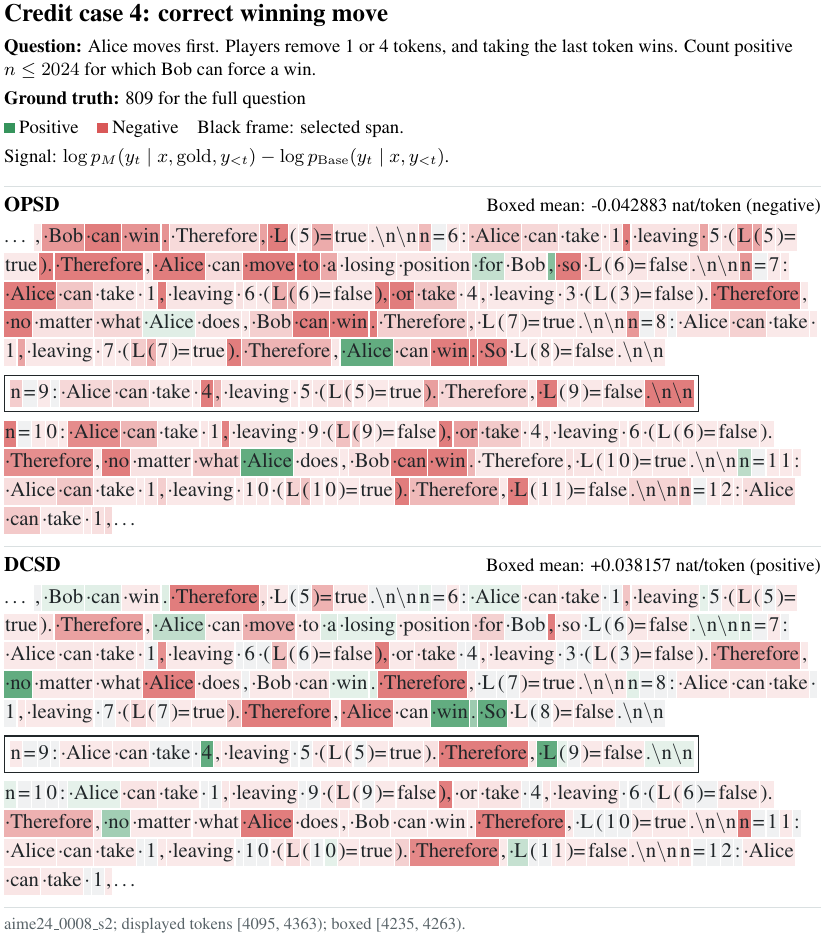}
\caption{\textbf{Correct winning move.} With nine tokens, taking four leaves five. Five is losing for the player to move: either allowed removal leaves the opponent a winning position. Thus nine is winning, consistent with the response's notation $L(5)=\mathrm{true}$ and $L(9)=\mathrm{false}$. The boxed inference warrants positive local credit. OPSD assigns a negative mean ($-0.042883$), whereas DCSD assigns a positive mean ($+0.038157$).}
\label{fig:credit-case-04}
\end{figure}\clearpage

\begin{figure}[!ht]
\centering
\includegraphics[width=\linewidth,height=0.80\textheight,keepaspectratio]{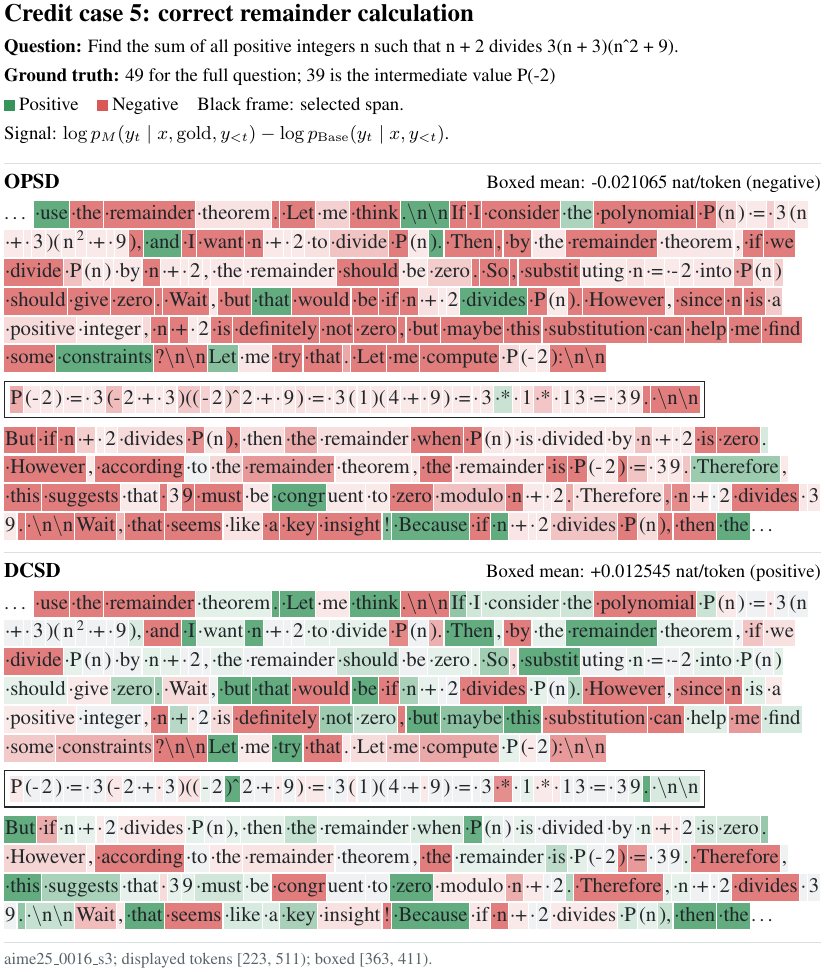}
\caption{\textbf{Correct remainder calculation.} The boxed substitution gives $P(-2)=3(-2+3)((-2)^2+9)=39$. Because $P(n)\equiv P(-2)\pmod{n+2}$, this calculation supplies the useful divisibility condition $n+2\mid39$; it is not a claim that $P(-2)$ must vanish. Positive local credit is appropriate for this intermediate step. OPSD assigns $-0.021065$, while DCSD assigns $+0.012545$, agreeing with the positive semantic judgment.}
\label{fig:credit-case-05}
\end{figure}\clearpage

\begin{figure}[!ht]
\centering
\includegraphics[width=\linewidth,height=0.80\textheight,keepaspectratio]{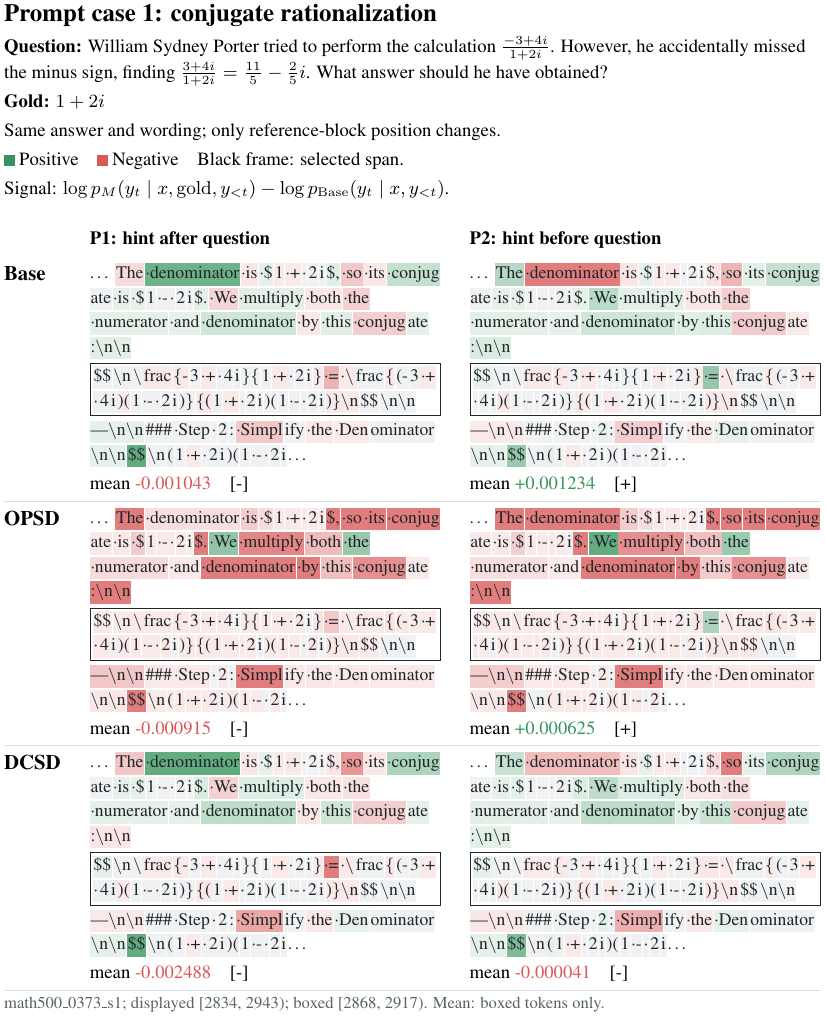}
\caption{\textbf{Correct conjugate rationalization.} Multiplying numerator and denominator by $1-2i$ is valid because the denominator is nonzero; it converts the denominator to $5$ and is a useful step deserving positive local credit. Base and OPSD change from small negative to small positive means across prompts. DCSD remains negative ($-0.002488$, $-0.000041$), with the second mean close to zero. This demonstrates sign stability, but not semantically correct positive credit for the boxed step.}
\label{fig:prompt-case-01}
\end{figure}\clearpage

\begin{figure}[!ht]
\centering
\includegraphics[width=\linewidth,height=0.80\textheight,keepaspectratio]{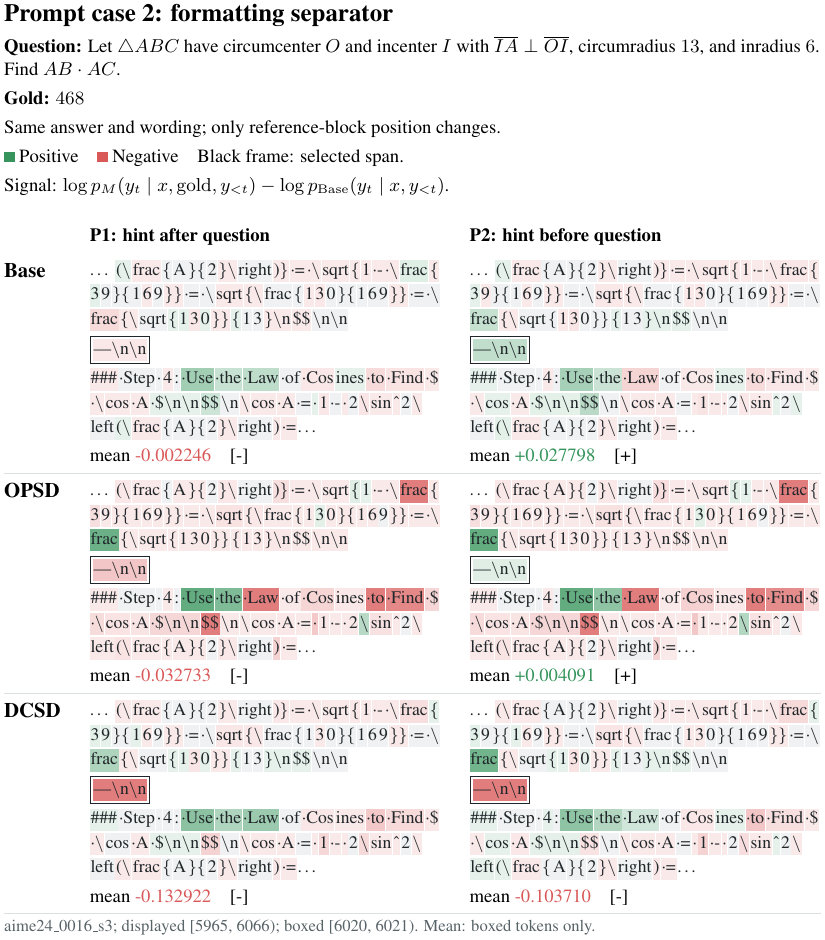}
\caption{\textbf{A formatting separator has no new mathematical content.} The black frame contains only a separator, not a reasoning assertion. For mathematical contribution its appropriate magnitude is near zero, without an intrinsically correct positive or negative sign. Base and OPSD switch from negative to positive across prompts. DCSD remains negative ($-0.132922$, $-0.103710$), showing stable suppression of this style token, but its substantial magnitude does not match a near-zero contribution target.}
\label{fig:prompt-case-02}
\end{figure}\clearpage

\begin{figure}[!ht]
\centering
\includegraphics[width=\linewidth,height=0.80\textheight,keepaspectratio]{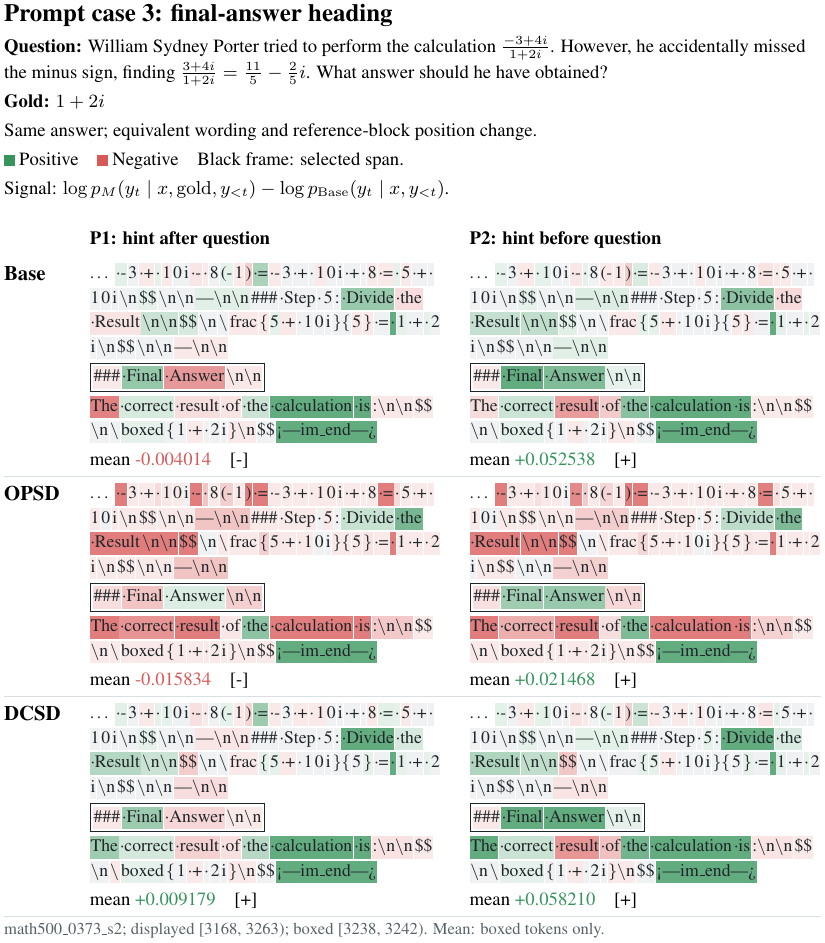}
\caption{\textbf{A final-answer heading is a stylistic cue.} The boxed text, ``Final Answer,'' adds no mathematical derivation; the actual answer $1+2i$ appears outside the frame. Its mathematical contribution should receive near-zero magnitude rather than a prescribed positive sign. Base and OPSD change from negative to positive, whereas DCSD remains positive ($+0.009179$, $+0.058210$). This case establishes direction stability across the selected prompts, not correctness of a positive mathematical-credit assignment.}
\label{fig:prompt-case-03}
\end{figure}\clearpage

\begin{figure}[!ht]
\centering
\includegraphics[width=\linewidth,height=0.80\textheight,keepaspectratio]{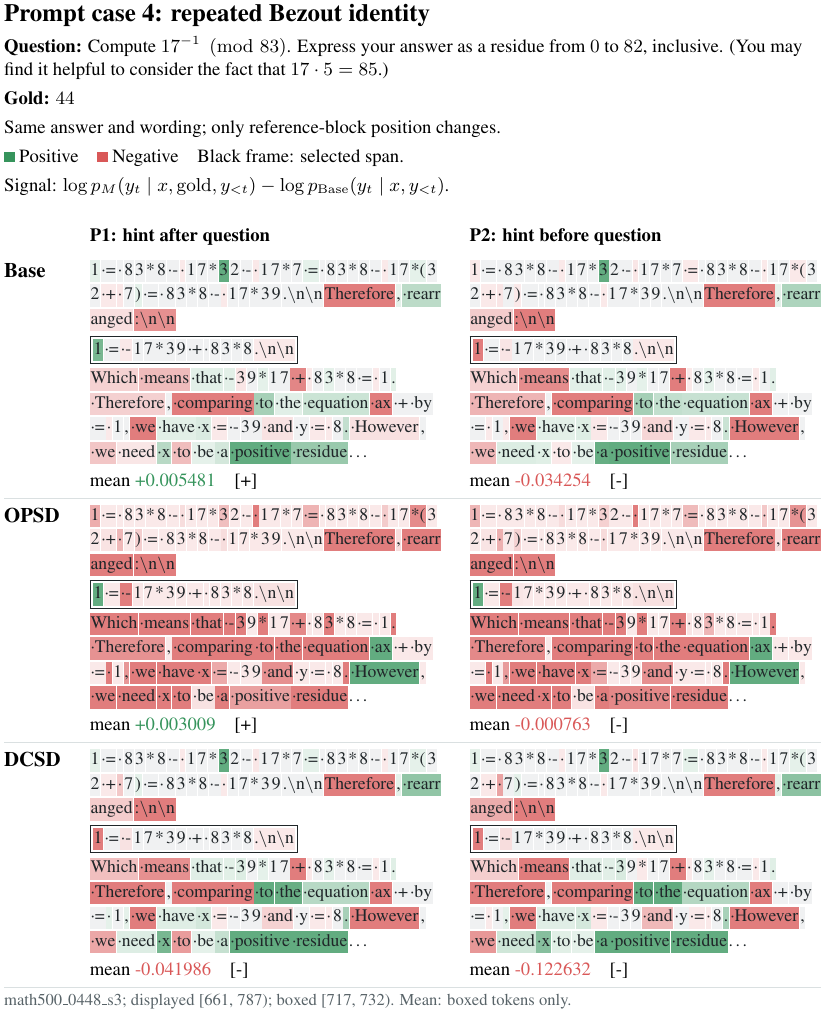}
\caption{\textbf{A correct but repeated B\'ezout identity.} The boxed equality $1=-17\cdot39+83\cdot8$ is true and yields the inverse $-39\equiv44\pmod{83}$. The preceding line already states the same identity, so this rearrangement has little additional mathematical content and merits low incremental magnitude. Base and OPSD change from positive to negative; DCSD remains negative ($-0.041986$, $-0.122632$). Its stable sign does not establish that the correct identity should receive negative credit.}
\label{fig:prompt-case-04}
\end{figure}\clearpage
\end{document}